\documentclass{article}
\usepackage{arxiv}

\usepackage{times}
\usepackage{soul}
\usepackage{url}
\usepackage[hidelinks]{hyperref}
\usepackage[utf8]{inputenc}
\usepackage[small]{caption}
\usepackage{booktabs}
\usepackage[title]{appendix}
\usepackage{enumerate}
\usepackage{float}
\usepackage{subcaption}
\usepackage{setspace}
\usepackage{romannum}
\usepackage{tabularx}
\usepackage{array}
\usepackage{mdframed}

\usepackage{amsmath,amsthm,amssymb,amsfonts,bm}
\allowdisplaybreaks
\usepackage{mathtools}
\usepackage{stackrel}                                   
\usepackage{nicefrac}                                   
\usepackage{accents}
\usepackage{scalerel}

\usepackage{algorithmic}
\usepackage[ruled,vlined,linesnumbered]{algorithm2e}    
\newcommand{\nonl}{\renewcommand{\nl}{\let\nl\oldnl}}   
\usepackage{listings}
\SetAlCapNameFnt{\footnotesize}
\SetAlCapFnt{\footnotesize}

\usepackage{dsfont}

\usepackage{graphicx}
\usepackage{longtable}
\usepackage{hhline}
\usepackage{makecell}
\usepackage{varwidth}                                   
\usepackage{wrapfig}                                    
\newcommand{\halfspace}{\kern 0.2em}
\usepackage[explicit,compact]{titlesec}                         

\usepackage[bottom]{footmisc}
    \newcommand{\trps}{^{\tiny \mathsf{T}}}										
    \newcommand{\dTV}[1]{\mathrm{d}_{\mathrm{TV}}\left(#1\right)}
    \DeclarePairedDelimiter\abs{\lvert}{\rvert}                 
    \DeclarePairedDelimiter\norm{\lVert}{\rVert}
    \renewcommand{\exp}[1]{\halfspace \mathrm{exp} \left[#1 \right]}
    
    \newcommand{\KL}{\text{\tiny KL}}
    \newcommand{\pl}{^\text{\tiny pl}}
    \newcommand{\op}{^\text{\tiny op}}
    \newcommand{\Nash}{\text{\tiny Nash}}
    \newcommand{\Learn}{\text{\tiny L}}
    
    \newcommand{\A}{\mathcal{A}}
    \newcommand{\B}{\mathcal{B}}
    \newcommand{\D}{\mathcal{D}}
    
    \newcommand{\V}{\mathcal{V}}
    \newcommand{\R}{\mathcal{R}}
    \newcommand{\E}{\mathbb{E}}
    \newcommand{\F}{\mathcal{F}}
    \newcommand{\Q}{\mathcal{Q}}
    \newcommand{\G}{\mathcal{G}}
    \newcommand{\T}{\mathcal{T}}
    \renewcommand{\S}{\mathcal{S}}
    \newcommand{\U}{\mathcal{U}}
    \newcommand{\W}{\mathcal{W}}
    \newcommand{\X}{\mathcal{X}}
    \renewcommand{\Re}{\mathbb{R}}
    
    \usepackage{thmtools}
    \usepackage{thm-restate}

    \newtheorem{theorem}{Theorem} 
    \newtheorem{lemma}[theorem]{Lemma}
    \newtheorem{assumption}{Assumption}
    
    \newtheorem{corollary}[theorem]{Corollary}

    \graphicspath{{./figures/}}             

\title{Learning Nash Equilibria in Zero-Sum Stochastic Games via Entropy-Regularized Policy Approximation}

\author{
    Yue Guan  \textsuperscript{1 *} \quad 
    Qifan Zhang \textsuperscript{2 *}\quad  
    Panagiotis Tsiotras \textsuperscript{2}\\
    \textsuperscript{1}
    Georgia Institute of Technology\\
    \textsuperscript{2}
    Mastercard Corporation\\
    \texttt{yguan44@gatech.edu, qifan.zhang@mastercard.com, tsiotras@gatech.edu}
}

\begin{document}

\maketitle
\begingroup\renewcommand\thefootnote{*}
\footnotetext{Equal contribution.}
\endgroup

\pagenumbering{arabic}
\begin{abstract}
We explore the use of policy approximations to reduce the computational cost of learning Nash equilibria in zero-sum stochastic games. 
We propose a new Q-learning type algorithm that uses a sequence of entropy-regularized soft policies to approximate the Nash policy during the Q-function updates. 
We prove that under certain conditions, by updating the entropy regularization, the algorithm converges to a Nash equilibrium. 
We also demonstrate the proposed algorithm's ability to transfer previous training experiences, enabling the agents to adapt quickly to new environments. 
We provide a dynamic hyper-parameter schedule scheme to further expedite convergence. 
Empirical results applied to a number of stochastic games verify that the proposed algorithm converges to a Nash equilibrium, while exhibiting a major speed-up over existing algorithms.
\end{abstract}

\section{Introduction}

Stochastic Games (SG)~\cite{owen:Game-Theory} is a widely adopted framework to extend reinforcement learning~\cite{Intro-ML:1998} to multiple-agent scenarios.
The resulting multi-agent reinforcement learning (MARL) framework assumes a group of autonomous agents that choose actions independently and interact with each other to reach an equilibrium~\cite{MARL-survey:Busoniu}.
When all agents are rational, the most natural solution concept is the one of a 
Nash Equilibrium (NE)~\cite{Nash:1951}.
The difficulty of extending single-agent RL methods to learn the NE stems from the fact that the interactions among agents result in a non-stationary environment and thus make learning computationally expensive and difficult to stabilize~\cite{MARL-learner-interactions:Matignon}.

Several approaches have been proposed to learn policies in multi-agent scenarios.
One set of algorithms learn directly a NE using computationally demanding operators,
such as Minimax-Q~\cite{MiniMax-Q:1994} and Nash-Q~\cite{Nash-Q:2003}.
Agents adopting these algorithms follow more rational policies but in doing so, they tend to take more time to learn these policies.
For example, Minimax-Q needs to solve a linear program to compute the Nash equilibrium at each Q-function update. 
Even with the help of neural networks, the number of linear programs solved during the learning process still grows with the cardinality of the state and action spaces, which makes Minimax-Q infeasible for large games.

Other popular approaches that extend single-agent RL methods~\cite{neuro-dynamic-programming} to multi-agent scenarios include: simplifying the interactions~\cite{LOLA:Foerster,Policy-Hill-Climb:Bowling,friends-or-foe:littman}, or introducing extra mechanisms in the game~\cite{MADDPG:2017,M3DDPG,foerster2017counterfactual}.
Such algorithms allow the agents 
to compute ``fast'' but less rational policies, but are unlikely to converge to a rational equilibrium, in general. 

The recent works on entropy-regularized soft Q-learning \cite{Fox:2015} 
provide an efficient way to approximate Nash policies. 
Specifically,
the two-agent Soft-Q algorithm \cite{Balance-Two-Player-Soft-Q:2018} avoids the use of the expensive linear optimizations to update the Q-function.
The two agents, instead, compute closed-form soft-optimal policies under an entropy regularization that explores the policy space close to the given priors.
Due to fixed regularization, however, the generated policies of the two-agent Soft-Q may be far from a NE.
Consequently, there is a need for algorithms that learn the NE, yet in a computationally efficient manner, that can be used in a wide variety of multi-agent situations. 

To achieve convergence to NE while maintaining computational efficiency, 
we propose a novel adaptive entropy-regularized Q-learning algorithm for zero-sum games, referred to as Soft Nash Q$^2$-learning (SNQ2L)
\footnote{SNQ2 refers to the non-learning version of the algorithm, which has access to the game dynamics.}. 
The proposed algorithm learns two different Q-values: a standard Q-value and an entropy-regularized soft Q-value.
The two values are used asynchronously in a ``feedback'' learning mechanism: 
the soft policies act as an efficient{
\parfillskip=0pt
\parskip=0pt
\par}
\begin{wrapfigure}{r}{0.3\textwidth}
\vspace{-3pt}
    \centering
    \includegraphics[width=0.8\linewidth]{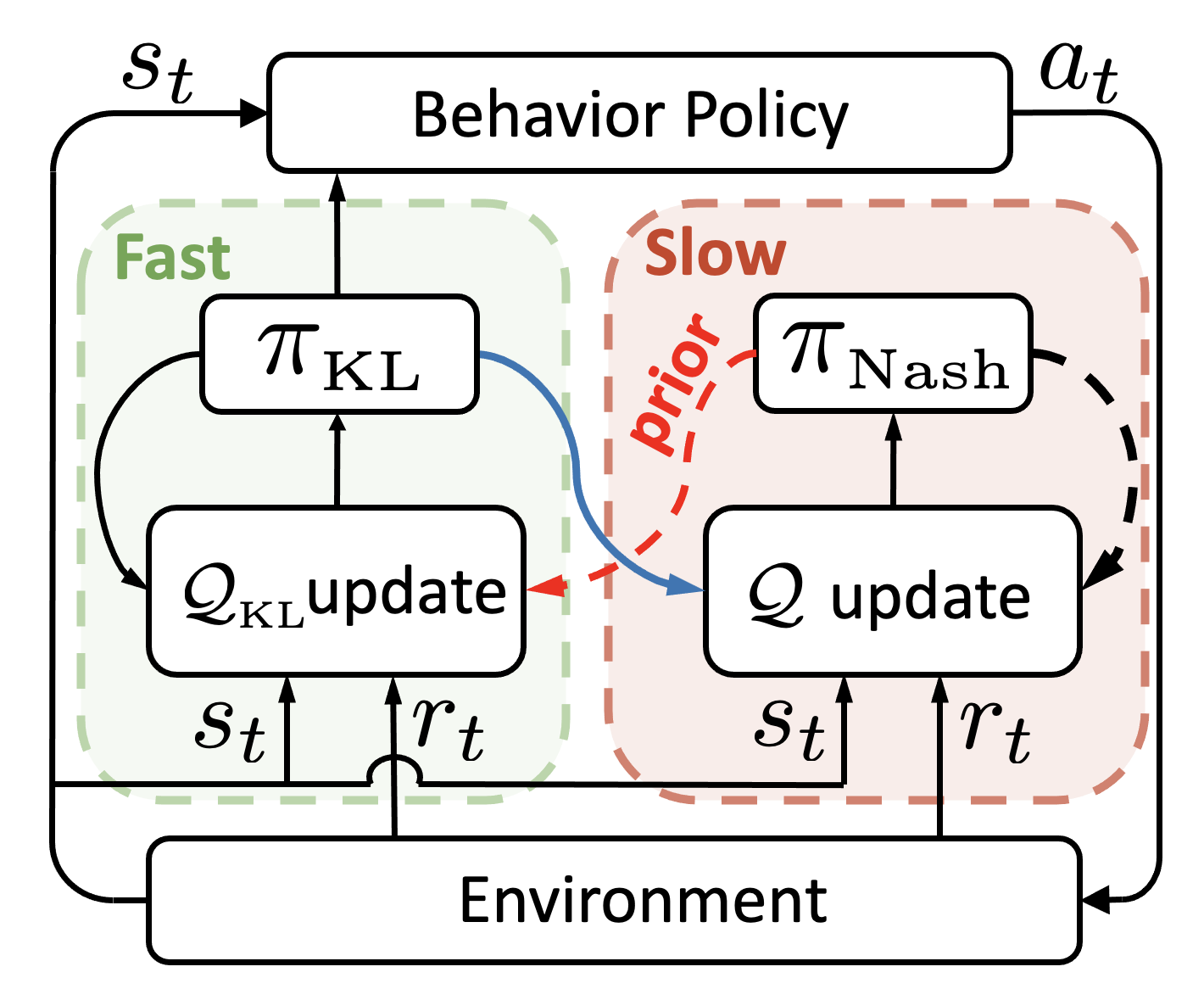}
    \vspace{-5pt}
    \caption{A schematic of SNQ2L.}
    \vspace{-20pt}
    \label{fig:SNQ2-Schemetic}
\end{wrapfigure}
\noindent approximation of the Nash policies to update the standard Q-value;
the Nash policies from the standard Q-function are computed periodically to update the priors that guide the soft policies (see Figure~\ref{fig:SNQ2-Schemetic}).
Consequently,
SNQ2L reduces the frequency of using the expensive Minimax operator, and thus expedites convergence to the NE.
Since the balance between the prior update frequency and computational efficiency plays a critical role in the performance of the SNQ2L algorithm,
we also introduce a dynamic schedule scheme that adaptively changes the prior update frequency.
The proposed algorithm has the potential of transferring previous experiences to new environments by incorporating a prior to ``warm start'' the learning process.

\paragraph{Contributions.}
The contributions of this paper can be summarized as:
(1) we propose a new algorithm for multi-agent games and prove the convergence of the proposed algorithm to a NE;
(2) we demonstrate a major speed-up in convergence to a NE over existing algorithms;
(3) we provide a dynamic schedule scheme of hyper-parameters to further expedite convergence;
and 
(4) we demonstrate the ability of the algorithm to transfer previous experience to a new environment.

\section{Background}\label{sec:background}
\paragraph{Two-agent Zero-sum Stochastic Games.}
In two-agent stochastic games, two agents, henceforth referred to as the Player (pl) and the Opponent (op), respectively, interact in the same stochastic environment. 
In this paper we are  interested, in particular, in zero-sum games where one of the agent's gain is the other agent's loss~\cite{cMDP:2012}. 
A zero-sum stochastic game $\G$ is formalized by the tuple $\G =\langle \mathcal{S}, \mathcal{A}\pl, \mathcal{A}\op, \mathcal{T},\mathcal{R},\gamma \rangle$, where
$\mathcal{S}$ denotes a finite state space, 
and $\mathcal{A}\pl$ and $\mathcal{A} \op$ are the finite action spaces.
To choose actions, the Player uses a (Markovian) policy $\pi\pl:S \times \mathcal{A}\pl \to [0,1]$ and the Opponent uses $\pi\op:S \times \mathcal{A}\op \to [0,1]$, which together produce the next state according to the state transition function $\mathcal{T}:\mathcal{S} \times \mathcal{S} \times \mathcal{A}\pl \times \mathcal{A}\op \to [0,1]$.
As a consequence of simultaneously taking these actions, the agents receive a reward $\R: \S \times \A\pl \times \A\op \to [R_{\min},R_{\max}]$.
The constant $\gamma \in (0,1)$ is the discount factor. 
For zero-sum games, the Player seeks to maximize the total expected reward, whereas the Opponent seeks to minimize it.
We denote the \textit{value} at each state induced by the policy pair $\pi=(\pi\pl,\pi\op)$ as $\V^\pi(s) = \E^{\pi}\left[\sum_{t=0}^\infty\gamma^t \R(s_t,a_t\pl,a_t\op) | s_0 = s\right]$. 
We denote by $\Pi$ the set of all admissible policy pairs.

\paragraph{Nash Equilibrium.}
Operating at a Nash equilibrium, no agent can gain by unilaterally deviating from her policy.
Formally, at a NE $(\pi^{\mathrm{pl}\star},\pi^{\mathrm{op}\star})$, 
at each state $s\in \S$ satisfies
\begin{equation*}
    \V^{\pi^{\mathrm{pl}},\pi^{\mathrm{op}\star}}(s) \leq 
    \V^{\pi^{\mathrm{pl}\star},\pi^{\mathrm{op}\star}}(s) \leq 
    \V^{\pi^{\mathrm{pl}\star},\pi^{\mathrm{op}}}(s),
\end{equation*}
for all admissible policies $\pi\pl,\pi\op$.
Even though multiple Nash equilibria could exist in a zero-sum SG, the minimax (optimal) value at  a NE is \textit{unique} and can be computed~\cite{vonNeumann:2007} via $\V^{*} (s)= \max_{\pi \pl} \min_{\pi \op} \V^{\pi \pl, \pi\op}(s) = \min_{\pi \op} \max_{\pi \pl} \V^{\pi \pl, \pi\op}(s)$.

\section{Sequential Policy Approximations}
The SNQ2L algorithm uses sequential policy approximations to relieve the computational burden seen in the Minimax-Q algorithm.
In this section, we introduce the baseline SNQ2 algorithm in a competitive MDP (cMDP) setting. 
That is, we assume the game $\G$ is known.
Later on, we will extend the algorithm for RL problems where $\G$ is only partially known.
The proposed algorithm uses entropy-regularized soft policies to approximate the Nash equilibrium at each iteration.
We prove the convergence of such algorithm to a NE in the Supplementary material.
In the process, we define four operators that are useful in presenting the proposed algorithm.

\vspace{+2pt}
\subsection{Shapley's Method}

Shapley's method~\cite{cMDP:2012} 
is an iterative algorithm commonly used to solve the Nash equilibrium of a zero-sum stochastic game
\footnote{Minimax-Q is an asynchronous learning version of Shapley's method.}.
Each iteration of the algorithm consists of two operations. 

\vspace{+2pt}
\paragraph{Step 1.} 
Compute the Nash policies at each state given the Q-function estimate at the current iteration $t$, by solving a linear program~\cite{cMDP:2012}. For example, the maximizing Player has the following optimization problem,
\begin{equation}\label{eqn:matrix-game-nash}
    \begin{alignedat}{2}
        &\quad \quad \max \quad                 && v \\ 
        & \text{subject to} \quad  && v\mathds{1}\trps - \bm{\pi}\pl(s) \trps \bm{\Q}_t(s) \leq 0\\
        &  \quad                    && \mathds{1}\trps \bm{\pi}\pl(s) = 1, 
        ~~ \bm{\pi}\pl(s) \geq 0,
    \end{alignedat}
\end{equation}
where $\bm{\pi}\pl(s)$ is the policy in vector form and $\bm{\Q}_t(s)$ is the Q-matrix at state $s$.
The Nash policy of the Opponent can be solved in a similar manner. 
We denote the solutions to these optimization problems as $\bm{\pi}\pl_{\Nash,t}(s)$ and $\bm{\pi}\op_{\Nash,t}(s)$, respectively.
Performing the optimizations at all states $s$ one may define the overall operation via the operator~$\Gamma_\Nash: Q \to \Pi$~as
\begin{equation}
    (\pi\pl_{\Nash,t},\pi\op_{\Nash,t}) = \Gamma_\Nash \halfspace \Q_t.
    \vspace{+4pt}
\end{equation}

\paragraph{Step 2.}
Given a policy pair $\pi$, update the Q-function via 
\begin{align}
    \Q_{t+1}&(s,a\pl,a\op) = \R(s,a\pl,a\op)  \label{eqn:standard-Q-update}\\
    &+ \gamma \sum_{s'\in\S}\T(s'|s,a\pl,a\op) \halfspace \bm{\pi}\pl(s')\trps \bm{\Q}_t(s') \bm{\pi}\op (s') .\nonumber
\end{align}
Similarly, we define the standard Q-function update operator as
$\Gamma_1: Q \times \Pi \to Q$. 
The Shapley's Method utilizes the following operator to iterate on the Q-functions.
\begin{equation}\label{eqn:Shapley-update-operator}
    \B(\Q) = \Gamma_1(\Q, \Gamma_\Nash\Q).
\end{equation}

Theorem~\ref{thm:shapley}~\cite{cMDP:2012} is the foundation for the convergence of both Shapley's method and Minimax-Q. 
\vspace{+6pt}
\begin{theorem}\label{thm:shapley}
    The operator $\B$ is a sup-norm contraction mapping with a contraction factor of $\gamma$.
    The fixed point of $\B$ is the Q-function corresponding to a Nash equilibrium.
\end{theorem}

\subsection{Entropy-Regularized Policy Approximation}
Entropy-regularized policy approximation~\cite{Fox:2015} was originally introduced to reduce the maximization bias commonly seen in learning algorithms.
This idea was later extended to two-agent scenarios and is referred to as the two-agent Soft-Q algorithm~\cite{Balance-Two-Player-Soft-Q:2018}.
Two-agent Soft-Q introduces two fixed entropy-regulation terms to the reward structure and thus restricts the policy exploration to a neighborhood of the given priors. 
Given a policy pair $(\pi\pl,\pi\op)$,
the soft-value function is defined as 
\begin{equation}\label{eqn:soft-Q-reward-structure}
        \V^{\pi\pl,\pi\op}_\KL(s) = \mathbb{E}_{s}^{\pi\pl,\pi\op} \bigg[ \sum_{t=0}^{\infty} \gamma^t \Big( \R(s_t,a_t\pl,a_t\op) \\
        -\frac{1}{\beta\pl}\log \frac{\pi\pl(a_t\pl|s_t)}{\rho\pl(a_t\pl|s_t)}
        -\frac{1}{\beta\op}\log \frac{\pi\op(a_t\op|s_t)}{\rho\op(a_t\op|s_t)} \Big)\bigg],
\end{equation}
where $\beta\pl>0$ and $\beta \op<0$ are inverse temperatures, and $\rho\pl$ and $\rho\op$ are priors for the Player and the Opponent.

The objective of the two-agent Soft-Q algorithm is to find the optimal solution to the max-min optimization problem 
$\max_{\pi\pl}\halfspace \min_{\pi\op} \V_\KL(s; {\pi\pl,\pi\op})$.
It has been shown in \cite{Balance-Two-Player-Soft-Q:2018} that the optimal solution can be found through an iterative algorithm.

\paragraph{Step 1.}
With the current soft Q-function $\Q_{\KL,t}$, construct the soft optimal policy for the Player through the closed-form solution 
\begin{equation}\label{eqn:soft-optimal-policies}
    \pi_{\KL,t}^{\text{pl}}(a\pl|s) 
    = \frac{1}{Z\pl(s)} \rho\pl(a\pl|s) \exp{ \beta\pl \Q_{\KL,t}^{\text{pl}}(s,a\pl)},
\end{equation} 
where $Z\pl(s)$ is the normalization factor and $\Q_{\KL,t}^{\text{pl}}$ is a marginalization through a log-sum-exp function
\begin{equation}\label{eqn:Q-marginalization}
    \Q_{\KL,t}\pl(s,a\pl) 
    =\frac{1}{\beta\op} \log \sum_{a\op} \rho\op(a\op|s) \mathrm{exp}\left[\beta\op \Q_{\KL,t}(s,a\pl,a\op)\right].
\end{equation}
The soft-optimal policy of the Opponent can be obtained in a similar manner. 
One can then define the soft optimal policy generating operator
$\Gamma_\KL^\beta: Q \times \Pi \to \Pi$ as
\begin{equation}
    (\pi\pl_{\Nash,t},\pi\op_{\Nash,t}) = \Gamma^\beta_\KL(\Q_{\KL,t}, \rho).
\end{equation}

\paragraph{Step 2.}
With the marginalization in \eqref{eqn:Q-marginalization}, compute the optimal value at state $s$ via 
\begin{equation}\label{eqn:soft-value}
    \V_{\KL,t}(s) = \frac{1}{\beta\pl} \log \sum_{a\pl} \rho \pl(a\pl|s) \exp{ \beta\pl \Q_{\KL,t}^\text{pl}(s,a\pl)}
\end{equation}

One can then update the Q-function through
\begin{equation}\label{eqn:soft-Q_update_MDP}
    \Q_{\KL,t+1}(s, a\pl,a\op)  
    = R(s,a\pl,a\op) + 
    \gamma \sum_{s'\in\S}\T(s'|s,a\pl,a\op) \V_{\KL,t}(s').
\end{equation}

We denote the soft update operator $\Gamma_2^\beta: Q \times \Pi \to Q$ as
\begin{equation}
    \Q_{\KL,t+1} = \Gamma^\beta_\KL(\Q_{\KL,t}, \rho).
\end{equation}
Note that, due to the fixed regularization, the two-agent Soft-Q algorithm does not converge to a NE in general.

\subsection{Sequential Policy Approximation Algorithm}
In Algorithm~\ref{alg:cMDP-SNQ2}, 
we present the baseline SNQ2 in the cMDP settings, where the game $\G$ is known.
Different from the two-agent Soft-Q with a fixed entropy regularization, the proposed algorithm decreases $\beta$ over time. 
As a result, a sequence of policy approximations of different levels of regularization is applied, and thus enables
the algorithm to converge to a NE. 

\setlength{\textfloatsep}{3pt}
\begin{algorithm}[t]
\SetAlgoLined
\caption{Baseline SNQ2 in the cMDP Settings}
\footnotesize
\label{alg:cMDP-SNQ2}
\lstset{numbers=left, numberstyle=\tiny, stepnumber=1, numbersep=5pt}
\footnotesize
\textbf{Inputs:} Game tuple $\G =\langle \mathcal{S}, \mathcal{A}\pl, \mathcal{A}\op, \mathcal{T},\mathcal{R},\gamma \rangle$\;
Set $\Q(s,a\pl,a\op) = \Q_{\KL}(s,a\pl,a\op) = 0$\; 
Set $\beta\pl$ and $\beta \op$ to some large values\; \label{baseline-Q2-Alg-init-end}
\While{$\Q$ and $\Q_\KL$ \text{ not converged}}{
    Update Nash policies: $\pi_\Nash \leftarrow \Gamma_\Nash(\Q)$; \\
    Update soft optimal policies: $\pi_\KL \leftarrow \Gamma^\beta_\KL(\Q_\KL,\pi_\Nash)$; \\
    Update Q-function: $\Q\text{\textunderscore} \leftarrow \Gamma_1(\Q, \pi_\KL)$; \\
    Update soft Q-function: $\Q_\KL\text{\textunderscore} \leftarrow \Gamma^\beta_2(\Q_\KL, \pi_\Nash)$; \\
    $\Q \leftarrow \Q\text{\textunderscore}$, $\Q_\KL \leftarrow \Q_\KL \text{\textunderscore}$; \\
    Reduce inverse temperature $\beta$
}
\Return $\pi_\Nash$ and $\Q(s,a\pl,a\op)$.
\end{algorithm}

Within the while loop, the algorithm first computes the Nash policies under the current estimate of the Q-function, and it uses the Nash policies as the priors to generate the soft policies. 
The soft policies are then used to update the Q-functions.

To prove the convergence of the baseline SNQ2 algorithm, we define the operations within the while loop in Algorithm~\ref{alg:cMDP-SNQ2} as $\Gamma^\beta: \bar{\Q} \to \bar{\Q}$, 
such that
\begin{equation*}
    \Gamma^\beta \bar{\Q} = \Gamma^\beta 
    \left[
\begin{array}{c}
     \Q \\ ~\Q_\KL
\end{array}
    \right]
    =
    \left[
\begin{array}{c}
     \Gamma_1(\Q,\Gamma_\KL^\beta (\Q_\KL, \Gamma_\Nash \Q))  \\
     \Gamma_2^\beta(\Q_\KL, \Gamma_\Nash \Q)
\end{array}
    \right].
\end{equation*}
The baseline SNQ2 algorithm differs from the standard iterative algorithms, in the sense that the operator $\Gamma^\beta$ changes at every iteration, as the inverse temperature pair $\beta = (\beta\pl,\beta\op)$ decreases to zero. 
As a result, we first present our theorem regarding the convergence of sequentially applying a family of contraction mappings. 
The proof of Theorem~\ref{thm:sequential-contraction} can be found in Appendix C.

\begin{theorem}\label{thm:sequential-contraction}
    Let $(\X,\rho)$ be a complete metric space, let $f_n: \X \to \X$ be a family of contraction mappings, 
    such that, for all $n=1,2,\ldots$ there exists $d_n\in(0,1)$, such that
    $\rho(f_n x, f_n y) \leq d_n \halfspace \rho(x, y)$ for all $x, y \in \X$.
    Assume that $\lim_{n \to \infty} d_n = d \in (0,1)$.
    Let $x \in \X$, and let $x^{(n)} = f_n\cdots f_1 x$ be the result of sequentially applying the operators $f_1, \ldots, f_n$ to $x$.
    If the sequence of operators $\{f_n\}_{n=1}^\infty$ converges pointwise to $f$, 
    then $f$ is also a $\rho$-contraction mapping with contraction factor $d$.
    Furthermore, if $x^\star$ is the fixed point of $f$,
    then, for every $x\in \X$, $\lim_{n \to \infty} x^{(n)} = x^\star$.
\end{theorem}
We now present the convergence theorem of Algorithm~\ref{alg:cMDP-SNQ2}. 
\begin{theorem}
    Under certain regularity assumptions, by sequentially applying $\Gamma^\beta$ and gradually decreasing $\beta$ to zero, Algorithm~\ref{alg:cMDP-SNQ2} converges to a Nash equilibrium. 
\end{theorem}
\begin{proof}
    We provide a sketch of the proof here. 
    The complete proof can be found at the supplementary material in Appendix C.
    We first show that under certain regularity assumptions, the operator $\Gamma^\beta$ is a family of contraction mapping.
    We then show that as $\beta$ approaches zero, the operator $\Gamma^\beta$ converges pointwise to an operator, which has the concatenated Nash Q-function $[\Q_\Nash^\star, \Q_\Nash^\star]\trps$ as its unique fixed point. 
    Then, by Theorem~\ref{thm:sequential-contraction}, we conclude that Algorithm~\ref{alg:cMDP-SNQ2} converges to a Nash equilibrium by sequentially applying $\Gamma^\beta$ to some initial Q-function and decreasing $\beta$ to zero.
\end{proof}
This baseline SNQ2 serves as a demonstration of the sequential policy approximation idea behind the SNQ2 algorithm. 
It does not relieve the computational burden of computing Nash policies, as we update the Nash policies at every iteration in the while loop. 
Due to the simplicity of the baseline algorithm, we can demonstrate its mechanism within the space limit.
The standard SNQ2, on the other hand, only applies $\Gamma_\Nash$ periodically. 
We present the standard SNQ2 algorithm and its convergence proof in the appendix.

\section{Two-Agent Soft Nash Q\texorpdfstring{$^2$}{TEXT}-Learning}\label{sec:algorithm}

In this section, we present the two-agent Soft Nash Q$^2$-Learning algorithm (SNQ2L).
This algorithm is an (asynchronous) learning version of the standard SNQ2 in Algorithm~\ref{alg:cMDP-SNQ2}. 
Specifically, this algorithm does not have access to the dynamics of the game and uses stochastic approximations to estimate the expectation operator in \eqref{eqn:soft-Q-reward-structure}.
This SNQ2L algorithm only updates the Nash policies periodically according to a dynamic schedule, 
and thus reduces the computations required to find the Nash equilibrium.

\paragraph{Learning Rule for $\Q_\KL$.}
\begin{equation}\label{eqn:soft-Q_update}
    \Q_{\KL,t+1}(s_t,a_t\pl,a_t\op) \leftarrow \left(1-\eta_t\right) \Q_{\KL,t}(s_t,a_t\pl,a_t\op)  
     +\eta_t \Big[\R(s_t,a_t\pl,a_t\op)+ \gamma \V_{\KL,t}(s_{t+1}) \Big].
\end{equation}
Here, $t$ is the learning step, and $\V_{\KL,t}(s_{t+1})$ is computed as in \eqref{eqn:soft-value} using the current $\Q_{\KL,t}$ estimate.
The priors used in \eqref{eqn:soft-value} are updated periodically using the Nash policies as indicated by the red arrow in Figure~\ref{fig:SNQ2-Schemetic}.

\paragraph{Learning Rule for $\Q$.}
\begin{equation}\label{eqn:Q_update}
    \Q_{t+1}(s_t,a_t\pl,a_t\op) \leftarrow \left(1-\alpha_{t}\right)\Q_{t}(s_t,a_t\pl,a_t\op) 
    + \alpha_{t} \left[ \R(s_t,a_t\pl,a_t\op) +\gamma \V_{t}(s_{t+1})  \right]. 
\end{equation}
To compute the estimated optimal value $\V_{t}(s_{t+1})$, SNQ2L uses either soft policies or Nash policies.
We summarize SNQ2L with dynamic schedule in Algorithm~\ref{alg:Soft_Q_Nash}. 

\begin{algorithm}[t]
\SetAlgoLined
\caption{SNQ2-Learning Algorithm}
\footnotesize
\label{alg:Soft_Q_Nash}
\lstset{numbers=left, numberstyle=\tiny, stepnumber=1, numbersep=5pt}
\footnotesize
\textbf{Inputs:} Priors $\rho$, Learning rates $\alpha$\ and $\eta$; 
initial prior update episode $M=\Delta M_0$;
Nash update frequency $T$\; \label{Q2-Alg-init-start} 
Set $\Q(s,a\pl,a\op) = \Q_{\KL}(s,a\pl,a\op) = 0$\;
Set $\beta\pl$ and $\beta \op$ to some large values\;
\While{$\Q$ \text{ not converged}}{
    \While{episode $i$ not end}{ \label{Q2-Alg-episode-start}
            Compute $\pi_\KL(s_t) \leftarrow \left[\Gamma^\beta_\KL(\Q_{\KL}, \rho)\right](s_t)$\;
            Collect transition $(s_t,a\pl_t,a\op_t,r_t,s_{t+1})$ where $a\pl_t\sim\bm{\pi}_\KL^{\text{pl}}(s_t)$,~  $a\op_t\sim \bm{\pi}_\KL^{\text{op}}(s_t)$\;
         \uIf {$t$ mod $T$ == 0} { \label{Q2-Alg-Nash-update-check}
            Compute 
            $ \mathcal{V}(s_{t+1}) = \max_{\pi\pl} \min_{a\op} ,
            \sum_{a\pl} \Q(s_{t+1},a\pl,a\op) \pi\pl \big(a\pl|s_{t+1} \big)$;
        }
        \Else{
            Compute 
            $\V(s_{t+1}) =  \bm{\pi}_{\KL}\pl (s_{t+1})\trps~ \bm{\Q}(s_{t+1}) ~\bm{\pi}_{\KL}\op(s_{t+1})$
            \;
        }
        Update $\Q(s_t,a\pl_t,a\op_t)$ with $\V(s_{t+1})$ via~\eqref{eqn:Q_update}\;
        Update $\Q_\KL(s_t,a\pl_t,a\op_t)$ via~(\ref{eqn:soft-Q_update})\;
    }
    \If {$i$ == M }{\label{Q2-Alg-prior-update-start}
        Compute $\pi_{\text{Nash}} \leftarrow \Gamma_\Nash \Q_t$\; 
        Update priors $\rho_{\text{new}}\leftarrow \pi_\Nash$\;
        Update schedule as in Algorithm~\ref{alg:dynamic-schedule}:\\
        \quad \quad $\Delta M, \beta_{\text{new}} = 
        \textit{DS}\big(\rho_{\text{new}},\rho, \beta,  \Delta M, \Q \big)$\;
        Update next prior update schedule $M$ += $\Delta M$\;
        Update priors $\rho \leftarrow \rho_{\text{new}}$, ~~ 
        $\beta \leftarrow \beta_{\text{new}}$\;
        Decrease learning rates $\alpha$ and $\eta$\;
    }\label{Q2-Alg-prior-update-end}
    
}
\Return $\Q(s,a\pl,a\op)$.
\end{algorithm}

\subsection{Schedule of the Hyper-parameters \texorpdfstring{$M$}{TEXT} and \texorpdfstring{$\beta$}{TEXT}}\label{subsec:dynamic-schedule}

As presented in Algorithm~\ref{alg:Soft_Q_Nash}, the trade-off between computational efficiency and approximation accuracy is embedded in the prior update frequency $\Delta M$ and the Nash update frequency $T$.
Intuitively, when the new Nash priors are close to the old ones, the algorithm is close to convergence.
In this situation, the prior update frequency should be decreased (increase $\Delta M$) and the algorithm should trust and exploit the priors (decrease $\beta$). 

The default number of episodes between two Nash prior policy updates $\Delta M_0$ and the default decay rate of the inverse temperature $\lambda \in (0,1)$ are given initially as
\begin{equation*}
    \scriptsize
    \Delta M_0 = \frac{N_\text{states} \times N_\text{action pairs}}{\alpha_0 \times T_\text{max}},
    \quad \quad \quad
    \lambda = \left(\frac{\beta_\text{end}}{\beta_0} \right)^{{1}/{N_{\text{updates}}}},
\end{equation*}
where $\alpha_0$ is the initial learning rate and $T_{\max}$ is the maximum length of a learning episode; $\beta_0$ and $\beta_\text{end}$ are the initial and estimated final magnitude for both
$\beta\op$ and $\beta\pl$, and $N_{\text{updates}}$ is the estimated number of prior updates.
This value of $\Delta M_0$ allows the algorithm to properly explore the state-action pairs so that the first prior update is performed with an informed Q-function. 
In our numerical experiments we found that $\beta_0 =20$, $\beta_\text{end} = 0.1$ and $N_{\text{updates}}=10$ are a good set of values. 

Algorithm~\ref{alg:dynamic-schedule} summarizes the dynamic schedule scheme, where the parameter $\sigma\in (0,1)$ is a decrease factor and the
$\textit{RelativeDifference}$ captures the performance difference between old and new priors.
We demonstrate the performance boost due to dynamic schedule in Section~\ref{subsec:numerical-ds}.

\begin{algorithm} [t]
\SetAlgoLined
\caption{Dynamic Schedule for $M$ and $\beta$}
\label{alg:dynamic-schedule}
\lstset{numbers=left, numberstyle=\tiny, stepnumber=1, numbersep=5pt}
\footnotesize
\textbf{Inputs:} old and new priors $\rho$, $\rho_{\text{new}}$;
old prior update length $\Delta M$;
old inverse temperatures $\beta$; 
current $\Q$\;
Compute ~ $\V_{\text{old}}(s) = [{\bm{\rho}\pl} (s)]\trps \bm{\Q}(s) \bm{\rho}\op(s)$
~ and ~
$\V_{\text{new}}(s) = {\bm{\rho}\pl_{\text{new}}} (s)\trps \bm{\Q}(s) \bm{\rho}\op_{\text{new}}(s)$, ~for all $s$\;
Compute ~$\textit{RelativeDifference}(s)
 = \abs{\V_{\text{new}}(s)-\V_{\text{old}}(s)}\big/\abs{\V_{\text{new}}(s)}$\;
Count the number of states $n$, where $\textit{RelativeDifference}(s)< \delta $\;
\uIf{${n}/{|\S|} \geq \text{Threshold}$}{
    $\Delta M = \min \{ \Delta M/\sigma, \Delta M_{\max} \}$,
    ~~
    $\beta_{\text{new}} = \max \{ \lambda \halfspace \beta_{\text{old}}, \beta_{\min}\} $\;
}
\Else{
    $\Delta M = \max \{\sigma \halfspace  \Delta M, \Delta M_{\min} \}$,
    ~~
    $\beta_{\text{new}} = \beta_\text{old} $\;
}
\Return $\Delta M$, $\beta_{\text{new}}$.
\end{algorithm}

\subsection{Warm-Starting}\label{subsec:warm-start}
One can warm-start the proposed SNQ2L algorithm by first initializing the priors $\rho \pl$ and $\rho \op$ based on some previously learnt policies or expert demonstrations, and then also postponing the first prior update to exploit these priors. 
Using a prior policy instead of the value to warm-start the algorithm has two major advantages: 
first, human demonstrations can be converted into prior policies in a more streamlined manner than into value information; 
second, prior policies provide more consistent guidance than the values, as the priors are not updated till the first prior update.  
We demonstrate the effectiveness of transferring previous experience in Section~\ref{subsec:numerical-warm-start}.


\section{Numerical Experiments}\label{sec:experiments}

To evaluate the performance of the proposed algorithm, we tested and compared SNQ2L
with four existing algorithms (Minimax-Q~\cite{MiniMax-Q:1994}, Two-agent Soft-Q~\cite{Balance-Two-Player-Soft-Q:2018}, WoLF-PHC~\cite{Policy-Hill-Climb:Bowling}, Single-Q~\cite{MARL:Tan}) for three zero-sum game 
environments:
a Soccer game as in~\cite{MiniMax-Q:1994}, a two-agent Pursuit-Evasion Game (PEG)~\cite{bounded-rational-PEG:2020} and
a sequential Rock-Paper-Scissor game (sRPS).

\subsection{Evaluation Criteria}\label{subsec:evaluation}
Two metrics were used to evaluate the performance of different algorithms:
the number of states achieving a NE and the running time%
\footnote{Implemented in a Python environment with AMD Ryzen 1920x.
Matrix games at each state are solved via Scipy's \texttt{linprog}.}.
For each game, we computed four different values and compared them to determine whether a state has achieved a NE:
(a) the ground truth Nash value $\V_{\Nash}$ solved exactly via Shapley's method; 
(b) the learnt value $\V_{\Learn}$; 
(c) the one-sided MDP value $\V_{\Learn}\pl$ computed by fixing the Opponent to her learnt policy and letting the Player maximize; 
(d) the one-sided MDP value $\V\op_\Learn$. 
We assume that the learnt policies achieve a NE at state $s$, if the values at state $s$ satisfy
\begin{equation*} 
   \max\Bigg\{\frac{\Big\lvert \V_{\Nash}(s)- \V_\Learn(s) \Big\rvert}{\big\lvert\V _ {\Nash}(s) \big\rvert},
   \frac{\Big\lvert \V _ {\Nash}(s) - \V_{\Learn}\pl(s) \Big\rvert}{\big\lvert\V _ {\Nash}(s) \big\rvert},
   \frac{\Big\lvert \V _ {\Nash} (s)- \V_{\Learn}\op (s)\Big\rvert}{\big\lvert\V _{\Nash}(s) \big\rvert}\Bigg\} < \epsilon.
\end{equation*}
Notice that the evaluation criterion above is stricter than the case of only collecting empirical win rates of different agents competing in the environment as in~\cite{MiniMax-Q:1994,value-approximation-SG}, since any deviation from the NE could be exploited by either agent in our criteria.

\subsection{The Game Environments} \label{subsec:game-envs}

\paragraph{Pursuit-Evasion.}

\begin{figure}[h]
    \vspace{-5pt}
    \centering
    \includegraphics[width = 0.5\linewidth]{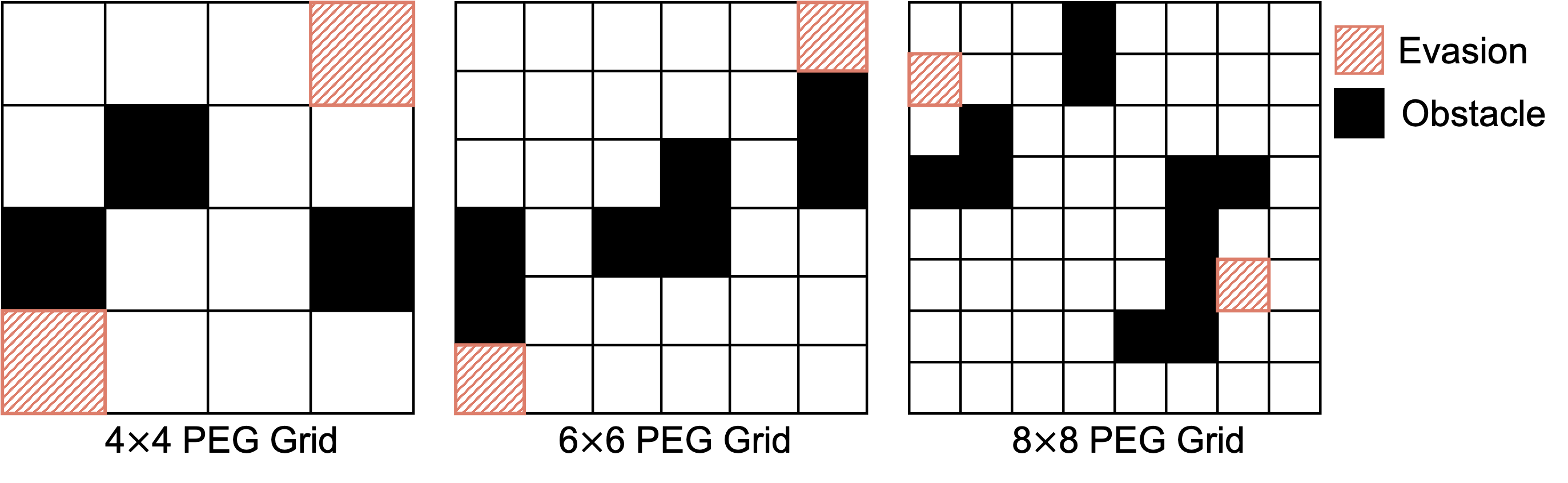}
    \vspace{-6pt}
    \caption{The grids of the three pursuit-evasion games.}
    \label{fig:PEG-Grid}
    \vspace{-0.5cm}
\end{figure}

The pursuit-evasion game (PEG) is played on 4$\tiny \times$4, 6$\tiny \times$6 and 8$\tiny \times$8 grids, as depicted in Figure~\ref{fig:PEG-Grid}. 
Both agents seek to avoid collision with the obstacles (marked in black).
The Pursuer strives to capture the Evader by being in the same cell as the Evader. 
The goal of the Evader is to reach one of the evasion cells (marked in red) without being captured. 
Two agents simultaneously choose one of four actions on each turn: \textit{N, S, E}, and \textit{W}.
Each executed action results in a stochastic transition of the agent's position.
In the default setting, an agent has a 60\% chance of successfully moving to its intended cell and a 40\% chance of landing in the cell to the left of the intended direction.
The 4$\tiny \times$4 and 8$\tiny \times$8 PEGs use the default transitions and the 6$\tiny \times$6 PEG uses deterministic transitions. 
In such setting, a deterministic policy cannot be a Nash policy.

\paragraph{Soccer.}

\begin{figure}[h]
\vspace{-5pt}
    \centering
    \includegraphics[width=0.6\linewidth]{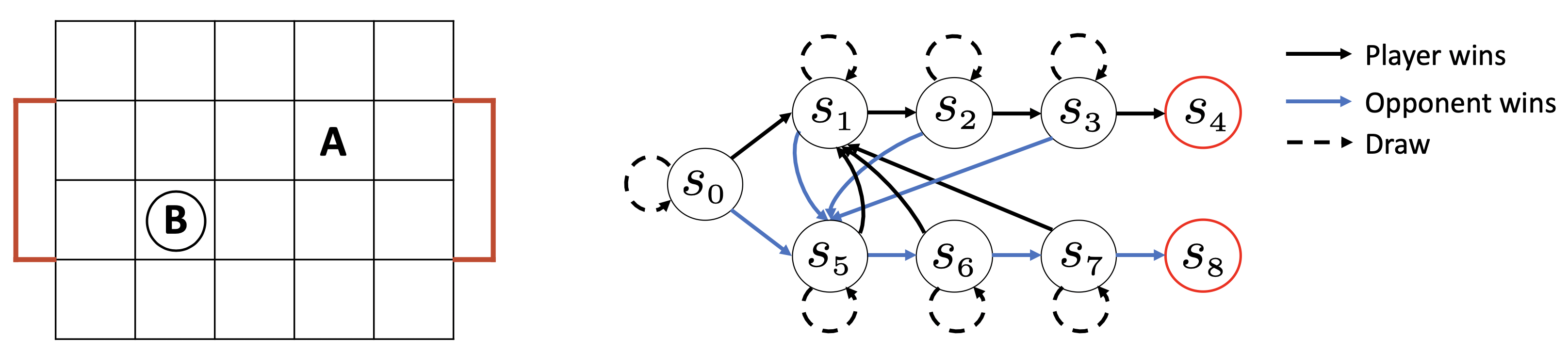}
    \vspace{-1pt}
    \caption{The grid of the soccer game (left) and the state transition diagram of sRPS (right). }
    \label{fig:soccer_sRPS}
    \vspace{0pt}
\end{figure}

The soccer game~\cite{MiniMax-Q:1994} is played on a 4$\tiny \times$5 grid as in Figure~\ref{fig:soccer_sRPS}. 
Two agents, A and B, can choose one of five actions on each turn: \textit{N, S, E, W,} and \textit{Stand}, and the two actions selected are executed in random order. 
The circle represents the ball. 
When the agent with the ball steps in to the goal (left for A and right for B), that player scores and the game restarts at a random state.
When an agent executes an action that would take it to the cell occupied by the other agent, possession of the ball goes to the stationary agent. 
Littman~\cite{MiniMax-Q:1994} argued that the Nash policies must be stochastic.

\paragraph{Sequential Rock-Paper-Scissor.}

In a sequential Rock-Paper-Scissor game (sRPS),
two agents ($\text{pl}$ and $\text{op}$) play Rock-Paper-Scissor repeatedly. 
One episode ends when one of the two agents wins four consecutive games.
The states and transitions are shown in Figure~\ref{fig:soccer_sRPS}. 
State $s_0$ corresponds to the initial state where no one has won a single RPS game, 
and states $s_4$ and $s_8$ are the winning (terminal) states of pl and op respectively.
The Nash policies of sPRS at each state are uniform for both agents.

\subsection{Comparison to Existing Algorithms}

\begin{figure*}[t]
\begin{subfigure}[b]{0.40\linewidth}
\centering
\includegraphics[width=\textwidth]{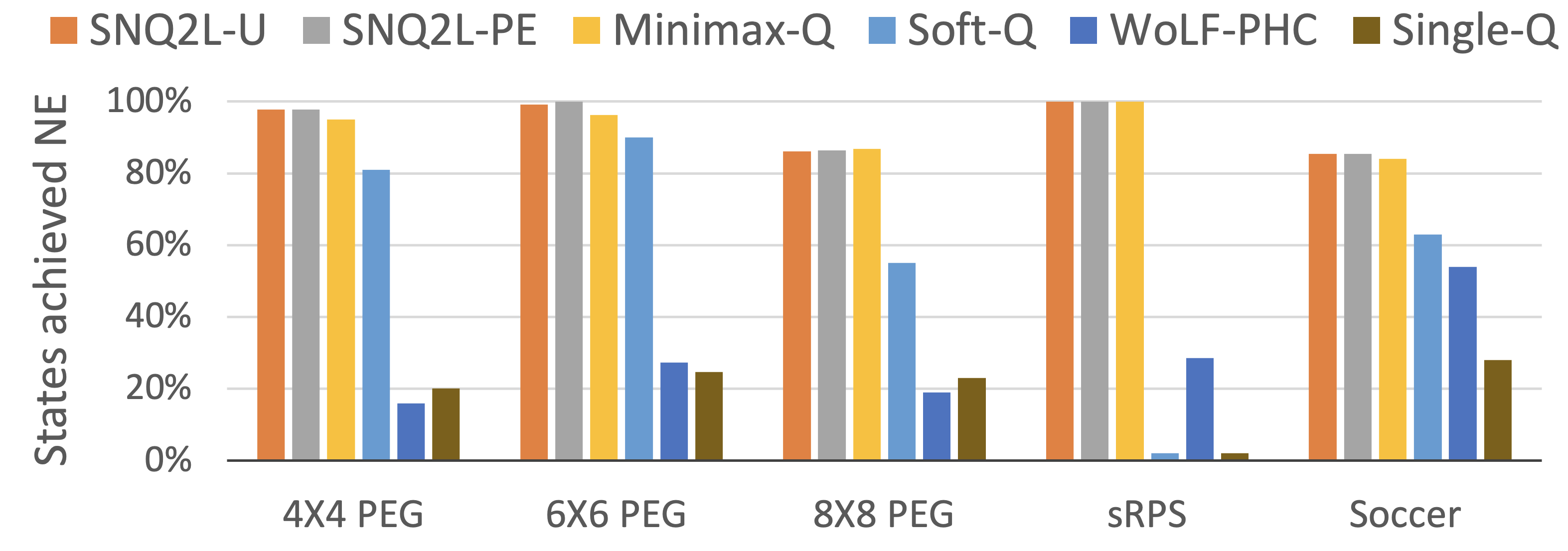}
\caption*{(a) Performance at convergence}
\end{subfigure}
\hspace{+3pt}
\begin{subfigure}[b]{0.24\linewidth}
\centering
\includegraphics[width=1\textwidth]{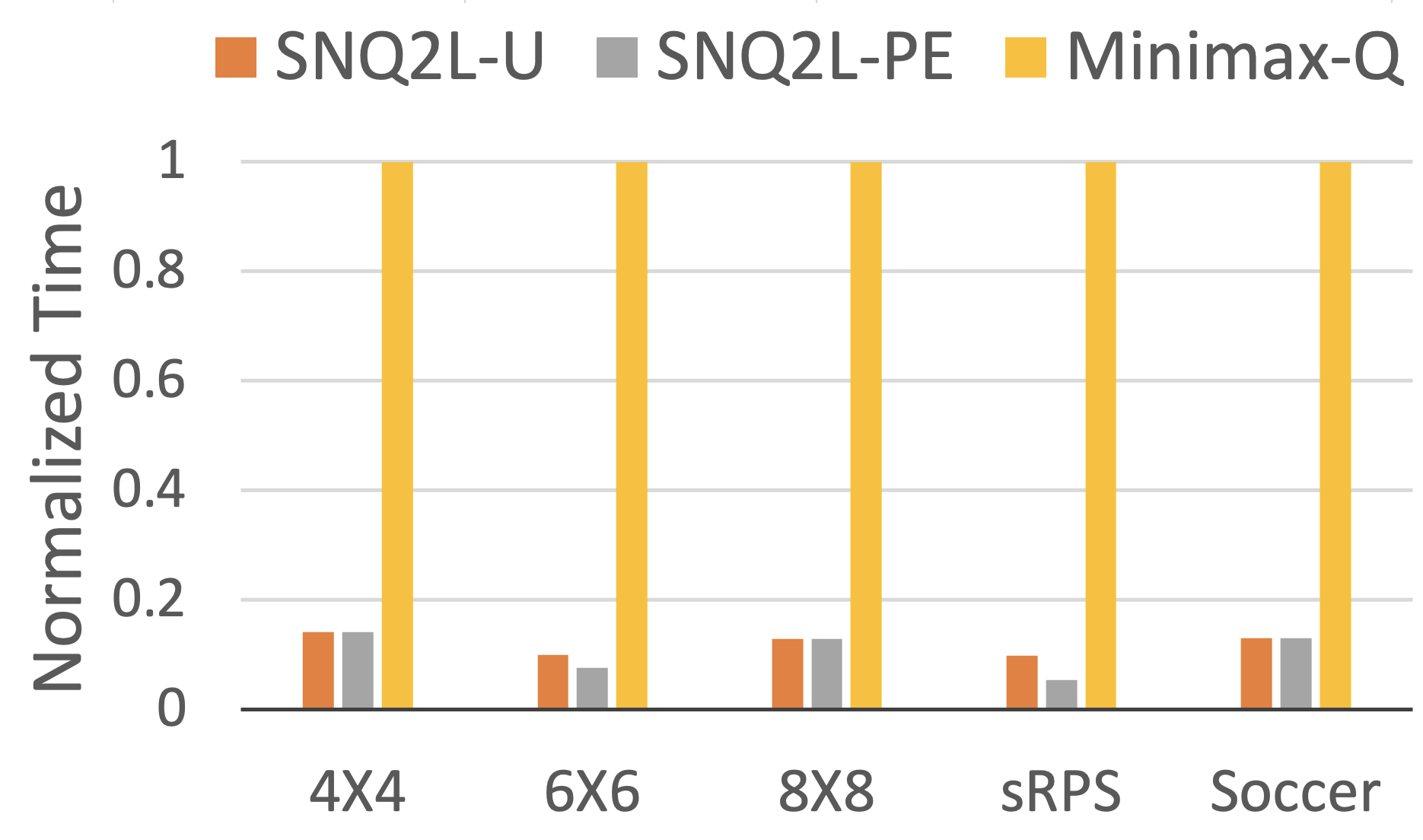}
\caption*{(b) Computation time}
\end{subfigure}
\hspace{+3pt}
\begin{subfigure}[b]{0.325\linewidth}
\centering
\includegraphics[width=1\textwidth]{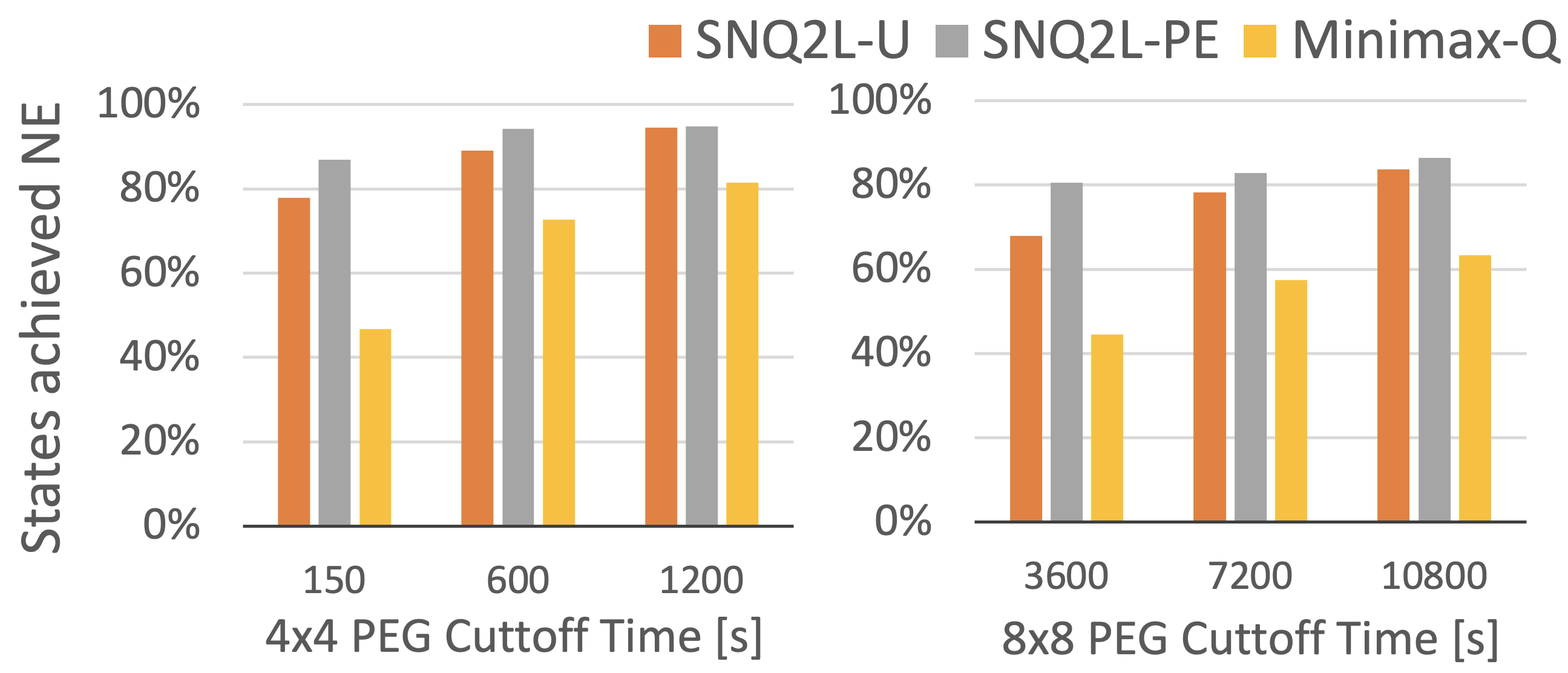}
\caption*{(c) Performance at cutoff }
\end{subfigure}
\vspace{-15pt}
\caption{
All results are averaged over ten runs.
The computation time in (b) is normalized by that of {Minimax-Q}.
We cut off the computation at 600k episodes for 8$\tiny \times$8 PEG, due to the large state space.
In most of the experiments, SNQ2L achieves a slightly better convergence to NE than {Minimax-Q}, while exhibiting an significant reduction in computation time. 
Two-agent Soft-Q, Single-Q and WoLF-PHC are fast but fail to converge to NE in all five games hence are not shown in (b).
Note that two-agent Soft-Q fails completely in sRPS.}
\vspace{+5pt}
\label{fig:performance-summary}
\end{figure*}

We evaluate SNQ2L
on the game environments in Section~\ref{subsec:game-envs} using the evaluation criteria in Section~\ref{subsec:evaluation}.
The SNQ2L algorithm can be initialized with two types of priors, uniform\footnote{For sRPS, the default prior is randomly generated, as uniform policies are the Nash policy for this game.}
(SNQ2L-U) and previous experience (SNQ2L-PE). 
Previous experience for PEGs and Soccer is learnt in a training session of the same game but with a different dynamics. 
For sRPS, the previous experience is a perturbed uniform strategy. 
The algorithm also has the option of a fixed schedule (SNQ2L-FS) and a dynamic schedule (SNQ2L-DS).
Unless otherwise specified, SNQ2L uses a dynamic schedule. 
SNQ2L was compared with four popular existing algorithms: Minimax-Q, two-agent Soft-Q, WoLF-PHC and Single-Q.
We fine-tuned these existing algorithms to get the best performance so as to demonstrate their actual capabilities.
We compare the convergence performance of the algorithms in Figure~\ref{fig:performance-summary}.

\setlength{\textfloatsep}{0pt}

The sRPS game with a uniform Nash equilibrium shows that SNQ2L is capable of converging to a fully mixed strategy.
In this game, two-agent Soft-Q learning fails to converge to the uniform NE, as its reward structure is regularized as in~\eqref{eqn:soft-Q-reward-structure};
Single-Q learning tries to learn a pure policy but does not converge;
WoLF-PHC converges to NE at two terminal states but with large value deviation, which propagates to other states and results in poor policies. 

\begin{figure}[h]
\begin{center}
    \begin{subfigure}[b]{0.4\linewidth}
\centering
\includegraphics[width=\textwidth]{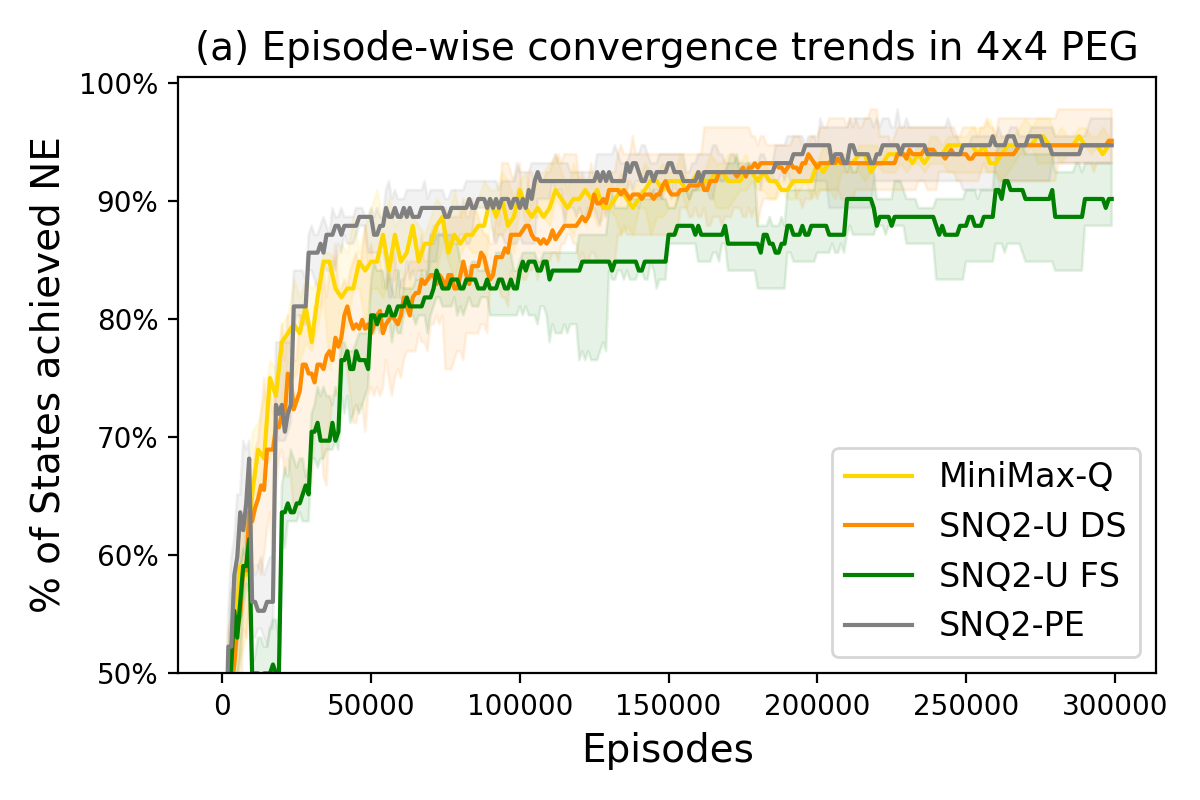}
\vspace{-13pt}
\end{subfigure}
\hspace{+20pt}
\begin{subfigure}[b]{0.4\linewidth}
\centering
\includegraphics[width=\textwidth]{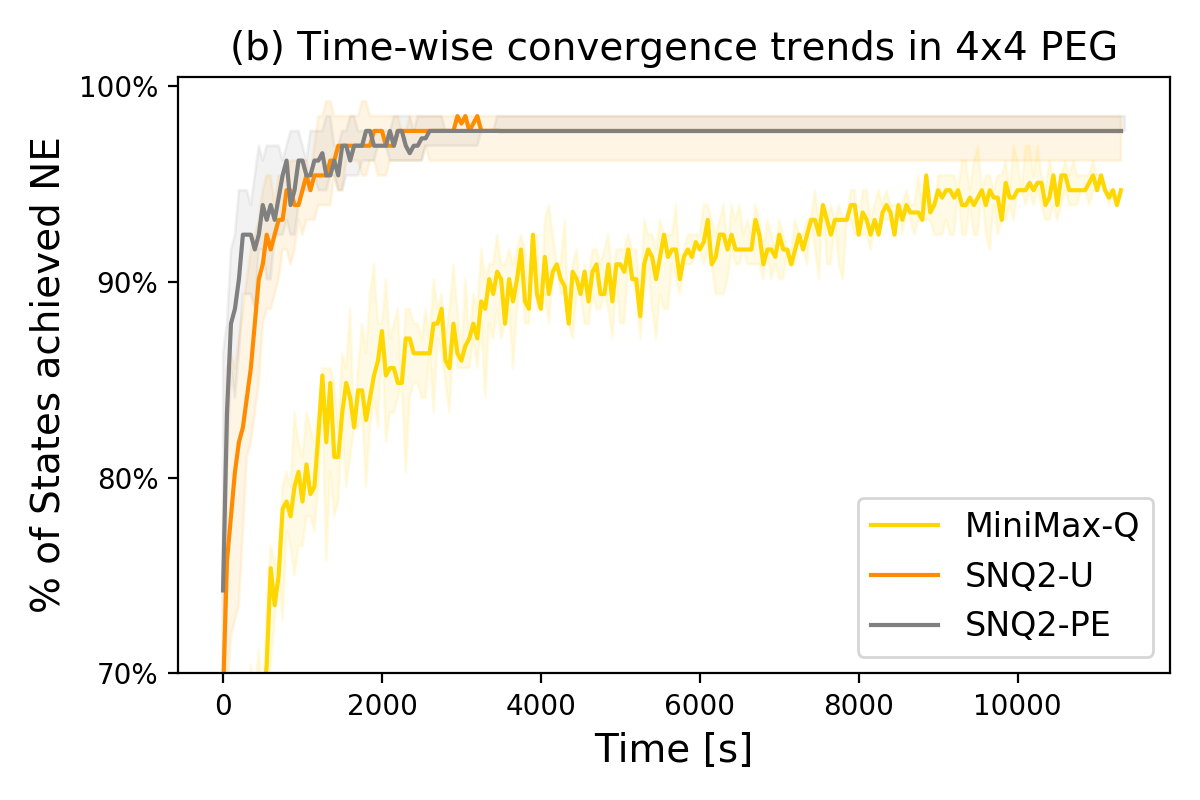}
\vspace{-15pt}
\end{subfigure}
\end{center}
\vspace{-5pt}
\caption{Convergence trends of different algorithms in 4$\tiny\times$4 PEG.}
\label{fig:4x4-conv}
\vspace{-5pt}
\end{figure}

We then examine the 4$\tiny \times$4 PEG game. 
Despite the relatively small state space, the stochastic transition at all non-terminal states requires extensive exploration of the environment.
We plot the convergence trends over 300k episodes and over 12k seconds for Minimax-Q and SNQ2L in Figure~\ref{fig:4x4-conv}.

The time-wise convergence trends in Figure~\ref{fig:4x4-conv} demonstrate the expedited convergence of SNQ2L in terms of computation time.
The episode-wise trend plot shows that SNQ2L maintains the same level of convergence to NE as Minimax-Q, albeit with significantly reduced computation time.
This shows that our adaptive entropy-regularized policy approximation approach is both accurate and efficient.

\subsection{Effectiveness of Warm-Starting}\label{subsec:numerical-warm-start}
In PEGs, the agents learn their previous experience on the same grid but with a 75\% success rate. 
To reduce overfitting to the previous environment, the prior policy fed to the agent for a new training session is the average of the previously learnt policy and a uniform policy. 
As seen in Figure~\ref{fig:4x4-conv}(b), previous experience does not shorten the time till convergence but instead significantly reduces the time to reach a reasonable performance.
We plot the cutoff time performance of the 4$\tiny \times$4 and 8$\tiny \times$8 PEGs in Figure~\ref{fig:performance-summary}(c).
In the 4$\tiny \times$4 example the time to reach over 90\% Nash convergence is halved from 1.2k seconds with uniform prior down to 600 seconds with previous experience.
In the 8$\tiny \times$8 example the time to 80\% convergence was halved from 7.2k seconds to 3.6k seconds.

One also observes from Figure~\ref{fig:4x4-conv}(a) that the policies generated by SNQ2L with previous experience converge, episode-wise, slightly faster than Minimax-Q. 
This ``warm-start'' feature has appealing real-world applications,
where the number of episodes interacting with the environment is the main constraint instead of computation time.
In this case, one can first train prior policies using a simulator and use these as priors to the agents. 
With SNQ2L one can train the agents to reach a reasonable level of performance in fewer episodes, while maintaining a relatively low computation overhead.


\subsection{Effectiveness of Dynamic Schedule} \label{subsec:numerical-ds}

\begin{figure}[t]
\begin{center}
    \begin{subfigure}[b]{0.4\linewidth}
\centering
\includegraphics[width=\textwidth]{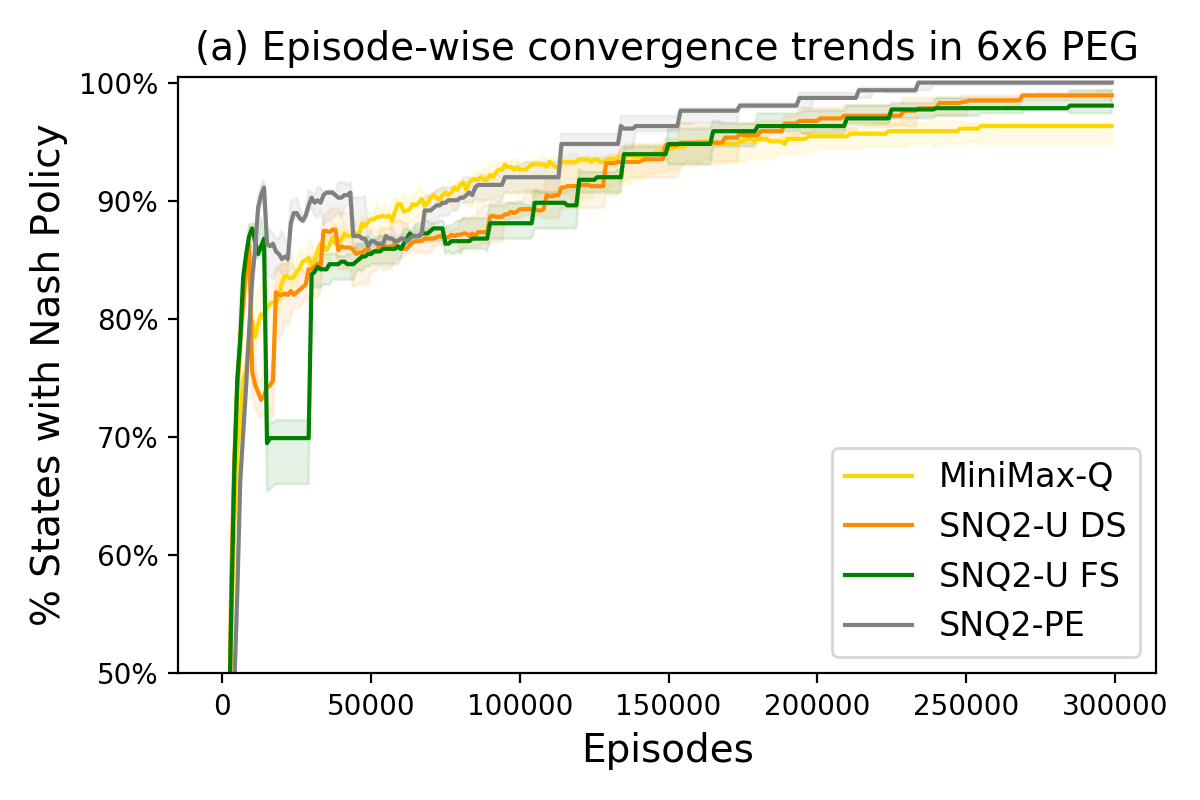}
\vspace{-13pt}
\end{subfigure}
\hspace{+20pt}
\begin{subfigure}[b]{0.4\linewidth}
\centering
\includegraphics[width=\textwidth]{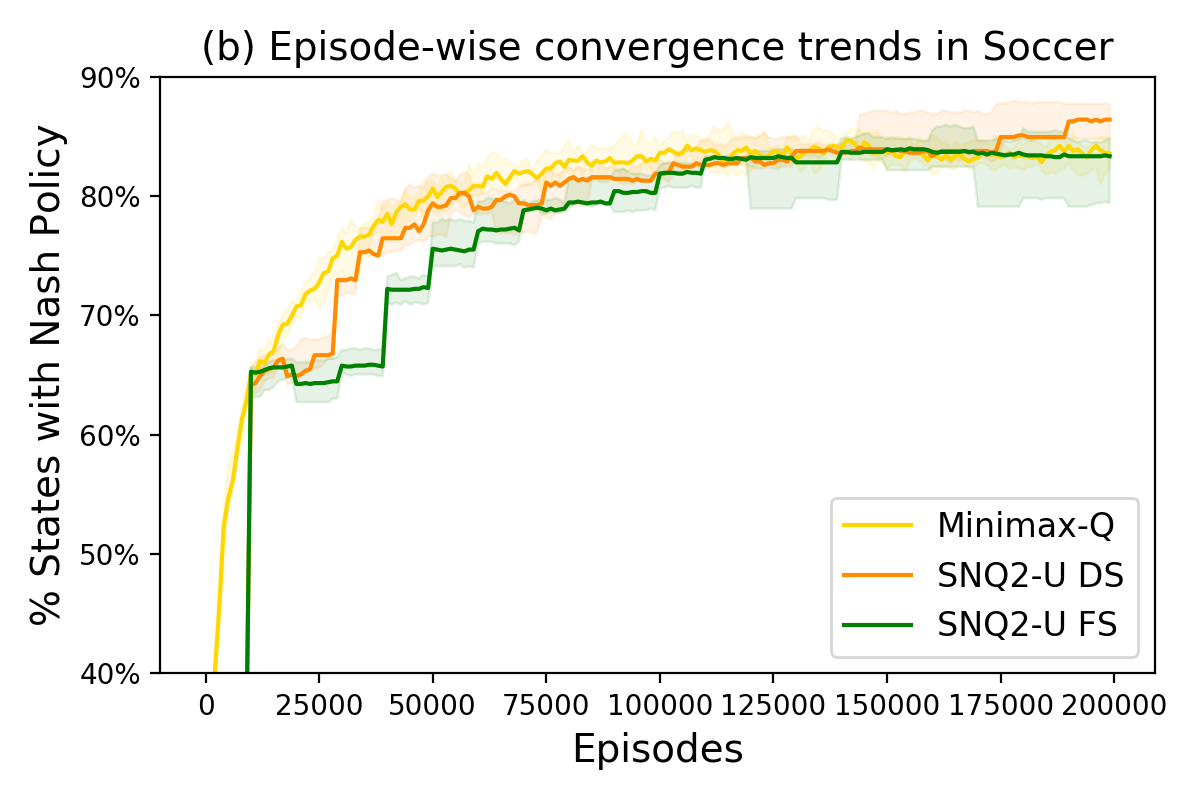}
\vspace{-13pt}
\end{subfigure}
\end{center}
\vspace{-5pt}
\caption{Episode-wise convergence trends in 6$\tiny\times$6 PEG and Soccer.}
\label{fig:dynamic-schedule-examples}
\vspace{+10pt}
\end{figure}

In Figure~\ref{fig:4x4-conv}(a), we use a fixed schedule (FS) with constant prior update intervals and we compare with the dynamic schedule (DS).
The latter demonstrates a faster episode-wise convergence
by reducing the number of episodes in which SNQ2L exploits a bad initial prior at the beginning of the learning.
As the initial learning rate is large, a Q-value based on a bad prior could be difficult to correct later on.
When the algorithm is close to convergence, 
DS reduces the prior update frequency and thus reduces performance oscillations, as shown in Figure~\ref{fig:4x4-conv}.
In Figure~\ref{fig:dynamic-schedule-examples}, the FS is fine-tuned through trial and error, and the DS is given a sub-optimal initial schedule. 
DS recovers from the given sub-optimal schedule and is still capable of achieving similar, or even better, performance compared to the hand-tuned FS.

\section{Conclusions and Future Work}\label{sec:conclusions}

We have proposed a novel algorithm for solving zero-sum games 
where the agents use entropy-regularized policies to approximate the Nash policies and thus reduce computation time till convergence.
We proved that under certain conditions, the proposed algorithm converges to a Nash equilibrium.
We conducted numerical experiments on multiple games of different sizes and levels of stochasticity.
The simulation results demonstrated that the algorithm significantly reduces computation time when compared to existing algorithms.
We also demonstrated the effectiveness of warm-starting in a new environment using previous experience. 
One interesting direction of future work is to extend the current algorithm to general-sum games.

\clearpage

\begin{appendices}
\vspace{-1.5cm}
\section{Derivation of the Two-Agent Soft-Q} \label{appd:soft-Q-derivation}

In this appendix, we derive the entropy-regulated policy approximations that are used in the SNQ2 algorithm. 
Most of the derivations in this section can be found in the existing literature (see for example \cite{Fox:2015} and \cite{Balance-Two-Player-Soft-Q:2018}).
The limiting case when the inverse temperature $\beta$ goes to zero is, however, only briefly discussed in the previous works, and without providing  rigorous justification.
As the convergence of the SNQ2 algorithm relies on the limiting behavior of the entropy-regulated policy approximation operators, we present several results to provide more insight to the limiting case where $\beta$ approaches zero.

We first revisit the policy approximation for the single-agent scenario (see~\cite{Fox:2015}).
We then utilize the single-agent results to derive the corresponding operators for the two-agent 
zero-sum game.

\subsection{Single-Agent Soft-Q Learning}

The single-agent entropy-regulated policy approximation is formulated as the following optimization problem~\cite{Fox:2015}
\begin{alignat} {2} 
    &\quad ~~ \max_{\pi} &&\mathbb{E}^\pi \left[\sum_{t=0}^{\infty} \gamma^t \R(s_t,a_t) \right], \label{eq100a}\\
    &\text{such that} \quad &&\mathbb{E}^\pi \left[ \sum_{t=0}^{\infty} \gamma^t \log \frac{\pi(a_t|s_t)}{\rho(a_t|s_t)}\right] \leq C, \label{eq100b}
\end{alignat}
where $\R(s,a)$ is the reward collected at state $s$ should the agent chooses action $a$ at that state.
The agent follows a policy $\pi$ and uses policy $\rho$ as its reference. 
The constraint in \eqref{eq100b} is meant to restrict exploration to policies that are close to the reference policy. 

The expectation operator in~(\ref{eq100a}) and~(\ref{eq100b})  is over all state-action trajectories ensuing by executing the policy $\pi$.
Notice that the expectation of the information cost over the policy $\pi$ in \eqref{eq100b} turns out to be the Kullback-Leibler (KL) divergence between $\bm{\pi}(s)$ and $\bm{\rho}(s)$. 
To solve  the optimization problem in~(\ref{eq100a})-(\ref{eq100b}) 
we introduce a Lagrangian multiplier ($-{1}/{\beta}$) to obtain an unconstrained optimization problem as follows
\begin{equation}
    \max_{\pi} ~ \mathbb{E}^\pi \left[\sum_{t=0}^{\infty} \gamma^t \R(s_t,a_t) \right] 
-    \frac{1}{\beta}
    \mathbb{E}^\pi \left[ \sum_{t=0}^{\infty} \gamma^t \log \frac{\pi(a_t|s_t)}{\rho(a_t|s_t)}\right].
\end{equation}

As a result of this Lagrangian relaxation, a positive inverse temperature $\beta$ corresponds to a maximization problem, and a negative $\beta$ corresponds to a minimization problem.

The magnitude of $\beta$ controls how far the optimal policy $\pi$ can deviate from the reference policy $\rho$.
If $\beta \to \infty$, the information cost constraint~(\ref{eq100b}) 
is no longer in effect, which can be seen from the complementary slackness condition.
If, instead, $\beta \to 0$, the optimal policy collapses to the reference policy.
This is formally proved later in this section. 
Intuitively, when $\beta = 0$, any deviation from the reference policy would induce an infinite information cost.

The policy $\pi$ that optimizes the above entropy-regulated optimization problem, is commonly referred to as the \textit{soft-optimal} policy. 
We will use the soft-optimal policies in the two-agent case to approximate the Nash policies in the next subsection.

Define the $\pi$-policy dependent soft value $\V^\pi_\KL$ as
\begin{equation}    \label{eqn-appd:derivation-end-1}
\begin{aligned} 
        \V^\pi_\KL(s) 
        &=  \mathbb{E}^\pi \left[\sum_{t=0}^{\infty} \gamma^t \left( \R(s_t,a_t) -\frac{1}{\beta} \log \frac{\pi(a_t|s_t)}{\rho(a_t|s_t)}\right) \Big\vert s_0 = s\right]   
         \nonumber \\
        &=\sum_{a \in \A} \pi(a|s)\left[ \R(s,a) -\frac{1}{\beta} \log \frac{\pi(a|s)}{\rho(a|s)} + \gamma \sum_{s'\in \S} \T_{ss' }(a)  \V^\pi_\KL(s')\right]\\
        &= \R(s,\bm{\pi}(s)) + \gamma \sum_{s'\in \S} \T_{s s'}\big(\bm{\pi}(s)\big)  \V^\pi_\KL(s')-\frac{1}{\beta} \mathrm{KL}\big(\bm{\pi}(s)||\bm{\rho}(s)\big),
\end{aligned}
\end{equation}
where we use the shorthand notation
\begin{align*}
    \R\big(s,\bm{\pi}(s)\big) = \sum_{a \in \A} \pi(a|s) \R(s,a), \\
    \T_{ss'}\big(\bm{\pi}(s)\big) = \sum_{a \in \A} \pi(a|s) \T_{ss'}(a).
\end{align*}
Next, we define the soft Q-function corresponding to the policy $\pi$ at state $s$ as
\begin{equation}   \label{eqn299:pt}
    \Q_{\KL}^{\pi}(s,a) = \R(s,a) + \gamma \sum_{s'\in \S} \T_{s s'}(a)  \V^\pi_\KL(s').
\end{equation}
Subsequently, we can also write
\begin{equation}   \label{eqn300:pt}
 \V^\pi_\KL(s) = \sum_{a \in \A} \pi(a|s)\Big(  \Q_{\KL}^{\pi}(s,a)
 -\frac{1}{\beta} \log \frac{\pi(a|s)}{\rho(a|s)} \Big).
\end{equation}

We define the soft-optimal value as
\begin{align}
    \V_\KL^\star(s) &= \max_{\pi} \V_\KL^\pi(s) \nonumber \\
    &= \max_{\bm{\pi}(s)} \sum_{a \in \A} \pi(a|s)\Big(  \Q_{\KL}^{\pi}(s,a)
 -\frac{1}{\beta} \log \frac{\pi(a|s)}{\rho(a|s)} \Big)\label{eqn-appd:v_KL}\\
    &= \max_{\bm{\pi}(s)} \Big(  \R(s,\bm{\pi}(s)) + \gamma \sum_{s'\in \S} \T_{s s'}\big(\bm{\pi}(s)\big)  \V^\pi_\KL(s')-\frac{1}{\beta} \mathrm{KL}(\bm{\pi}(s)||\bm{\rho}(s))  \Big).
\end{align}

The corresponding soft-optimal Q-function is then defined as
\begin{equation*}
    \Q_{\KL}^\star(s,a) = \R(s,a) + \gamma \sum_{s'\in \S} \T_{s s'}(a)  \V^\star_\KL(s').
\end{equation*}
Given a value vector $\halfspace \U \in \Re^{|\S|}$, 
we define the operator $\B:\Re^{|\S|} \to \Re^{|\S|}$
as follows
\begin{equation}\label{eqn-appd:single-agent-bellman}
    \big(\mathcal{B}\mathcal{U}\big)(s) 
    = \max_{\bm{\pi}(s)} \Big( \R(s,\bm{\pi}(s)) + \gamma \sum_{s'\in \S} \T_{s s'}\big(\bm{\pi}(s)\big)  \mathcal{U}(s')-\frac{1}{\beta} \mathrm{KL}(\bm{\pi}(s)||\bm{\rho}(s)) \Big).
\end{equation}
It follows that the optimal soft value $\V_\KL^*$ is a fixed point of the operator $\B$. That is, the optimization in~\eqref{eqn-appd:v_KL} reduces to finding the solution to the following equation
\begin{equation}
    \V_\KL^\star = \B \V_\KL^\star.
\end{equation}

\begin{lemma} \label{lemma-contraction}
The operator $\B$ in~(\ref{eqn-appd:single-agent-bellman}) is a sup-norm contraction with contraction factor $\gamma$.
\end{lemma}

\begin{proof}
    The sketch of the proof is as follows:
    One first defines the following evaluation operator $\B^\pi$ 
    \begin{equation}
    \big(\mathcal{B}^\pi \mathcal{U}\big)(s) 
    =\R(s,\bm{\pi}(s)) + \gamma \sum_{s'\in \S} \T_{s s'}\big(\bm{\pi}(s)\big)  \mathcal{U}(s')-\frac{1}{\beta} \mathrm{KL}(\bm{\pi}(s)||\bm{\rho}(s)).
\end{equation}
  and shows that it is a contraction mapping.
    One can then show that, with an extra max operator added to $\B^\pi$, the soft value update operator $\B$ is still a contraction mapping.
    This step is similar to proving  that the standard Bellman operator is a contraction mapping~\cite{neuro-dynamic-programming}.
    Refer to \cite{Fox:2015} and \cite{Balance-Two-Player-Soft-Q:2018} for more details regarding the proof.
\end{proof}

A major consequence of Lemma~\ref{lemma-contraction} is that one can find the soft optimal value $\V_\KL^\star$ via iterating the operator $\B$.
In the sequel, we use $\V_{\KL,t}$ to denote the estimate of the soft-optimal value at iteration $t$.
That is,
\begin{equation}
    \V_{\KL,t} = \B \V_{\KL,t-1}, \quad t=0,1,2,\ldots.
\end{equation}
Due to the contraction property of $\B$, we have
\begin{equation*}
    \V_\KL^\star = \lim_{t\to\infty} \V_{\KL,t} \halfspace.
\end{equation*}
The resulting estimate of the soft-optimal Q-function at iteration $t$, is then defined as
\begin{equation}\label{appd-eqn:single_Q_KL_t}
    \Q_{\KL,t}(s,a) = \R(s,a) + \gamma \sum_{s'\in \S} \T_{s s'}(a)  \V_{\KL,t}(s').
\end{equation}

Given the current estimate $\Q_{\KL,t}$ of the soft-optimal Q-function, the optimization problem in~\eqref{eqn-appd:v_KL} becomes 
\begin{align}
    \V_{\KL,t+1}(s) = 
    (\B \V_{\KL,t})(s) 
    = \max_{\bm{\pi}(s)} \sum_{a \in \A} \pi(a|s)\Big(\Q_{\KL,t}(s,a)
 -\frac{1}{\beta} \log \frac{\pi(a|s)}{\rho(a|s)} \Big)\label{eqn-appd:v_KL_t}.
\end{align}

To further obtain the soft-optimal policy that achieves the maximization problem defined in \eqref{eqn-appd:v_KL_t}, we need the following lemma from \cite{Fox:2015}.

\begin{lemma}\label{lemma:soft-optimal-policy}
    For each state $s \in \S$, the soft-optimal policy that solves the maximization problem~(\ref{eqn-appd:v_KL_t}) is given by
    \begin{equation}      \label{eqn-appd:single-agent-soft-optimal-policy}
        \pi_{\KL,t}(a|s) = \frac{\rho(a|s)\exp{\beta \Q_{\KL,t}(s,a)}}{\sum_{a'\in \A}\rho(a'|s)\exp{\beta \Q_{\KL,t}(s,a')}}.
    \end{equation}
\end{lemma}

\begin{proof}
    The optimization problem in \eqref{eqn-appd:v_KL_t} can be solved via a Lagrangian relaxation, and
    the solution to the optimization is \eqref{eqn-appd:single-agent-soft-optimal-policy}.
    Refer to \cite{Balance-Two-Player-Soft-Q:2018} for more details. 
\end{proof}

By substituting the soft-optimal policy $\pi_{\KL,t}$ from \eqref{eqn-appd:single-agent-soft-optimal-policy}
back into~\eqref{eqn-appd:v_KL_t}, one obtains a closed-form formula for the resulting soft-optimal value.

\begin{corollary}\label{cor:max-soft-optimal-value}
    Let $\Q_{\KL,t}$ be the estimate of the soft-optimal $Q$-function at iteration $t$. 
    Then, at each state $s \in \S$, the soft-optimal value corresponding to policy~\eqref{eqn-appd:single-agent-soft-optimal-policy} 
    is given by 
    \begin{equation}\label{eqn-appd:single-agent-soft-optimal-value}
    \begin{aligned}
        \V_{\KL,t+1}(s) &= \frac{1}{\beta} \log \sum_{a\in \A}\rho(a|s)\exp{\beta \Q_{\KL,t}(s,a)},
    \end{aligned}
    \end{equation} 
    where $\Q_{\KL,t}$ is defined through $\V_{\KL,t}$ as in \eqref{appd-eqn:single_Q_KL_t}.
\end{corollary}

A similar derivation works for the minimization version of the problem in~\eqref{eqn-appd:v_KL_t}. 
Hence, we have the following corollary.

\begin{corollary}\label{cor:min-optimal-policy-value}
Let $\beta<0$, and consider the following problem
\begin{equation}\label{eqn-appd:min-soft-optimal-problem}
    \V_{\KL,t+1}(s)
    = \min_{\bm{\pi}(s)} \sum_{a \in \A} \pi(a|s)\left( \Q_{\KL,t}(s,a)-\frac{1}{\beta} \log \frac{\pi(a|s)}{\rho(a|s)}\right).
\end{equation}
For each state $s\in\S$, the soft-optimal policy that solves the minimization problem (\ref{eqn-appd:min-soft-optimal-problem}) has the solution
\begin{equation*}
    \pi_{\KL,t}(a|s) = \frac{\rho(a|s)\exp{\beta \Q_{\KL,t}(s,a)}}{\sum_{a'\in \A}\rho(a'|s)\exp{\beta \Q_{\KL,t}(s,a')}},
\end{equation*}
and the soft-optimal value at each state $s\in\S$ is given by 
    \begin{equation}
        \V_{\KL,t+1}(s) = \frac{1}{\beta} \log \sum_{a\in \A}\rho(a|s)\exp{\beta \Q_{\KL,t}(s,a)}.
    \end{equation} 
\end{corollary}

Previous works such as \cite{Fox:2015} and \cite{Balance-Two-Player-Soft-Q:2018} pay 
little attention to the limiting case of the operator $\B$ as the inverse temperature approaches zero.
In our work, however, the limiting case of $\beta \to 0$ plays an essential role in the convergence of the proposed SNQ2 algorithm. 
As a result, we prove the following lemma regarding the soft optimal value when $\beta \to 0$.
\begin{lemma}\label{lemma:single-agent-limiting-value}
As the inverse temperature approaches zero, $\beta \to 0$, 
we have, for every $t=0,1,2,\ldots,$ that 
    \begin{equation*}
        \lim_{\beta \to 0} \V_{\KL,t+1}(s) = \sum_{a \in \A} \rho(s,a) \Q_{\KL,t}(s,a).
    \end{equation*}
\end{lemma}
\begin{proof}
Using L'Hospital's rule, we have, at each state $s$, 
\begin{align*}
    \lim_{\beta \to 0}  \V_{\KL,t+1}(s) &= \lim_{\beta \to 0} \frac{ \log \sum_{a\in \A}\rho(a|s)\exp{\beta \Q_{\KL,t}(s,a)}}{\beta}
    =\lim_{\beta \to 0} \frac{ \frac{\partial}{\partial \beta} \halfspace \log \sum_{a\in \A}\rho(a|s)\exp{\beta \Q_{\KL,t}(s,a)}}{\frac{\partial}{\partial \beta} \halfspace \beta}\\
    &= \lim_{\beta \to 0} 
    \frac{1}{\sum_{a\in \A}\rho(a|s)\exp{\beta \Q_{\KL,t}(s,a)}} \sum_{a\in \A}\rho(a|s) \Q_{\KL,t}(s,a) \exp{\beta \Q_{\KL,t}(s,a)}\\
    &=\sum_{a \in \A} \rho(s,a) \Q_{\KL,t}(s,a),
\end{align*}
where we have used the fact that $\sum_{a'\in\A}\rho(a'|s) = 1$ for all $s \in \S$.
\end{proof}

We conclude this subsection with the following lemma dealing with the soft-optimal policy as the inverse temperature approaches zero.

\begin{lemma} \label{lemma:single-agent-limiting-policy}
    As $\beta \to 0$, we have
    \begin{equation}
     \lim_{\beta \to 0}   \pi_{\KL,t}(a|s) = \rho(a|s).
    \end{equation}
\end{lemma}
\begin{proof}
The result follows directly from
    \begin{equation*}
        \begin{aligned}
            \lim_{\beta\to 0}\pi_{\KL,t}(a|s) &= \lim_{\beta\to 0}\frac{\rho(a|s)\exp{\beta \Q_{\KL,t}(s,a)}}{\sum_{a'\in \A}\rho(a'|s)\exp{\beta \Q_{\KL,t}(s,a')}}\\
            &= ~ \frac{\rho(a|s)}{\sum_{a'\in\A}\rho(a'|s)} = \rho(a|s).
        \end{aligned}
    \end{equation*}

\end{proof}

Lemma~\ref{lemma:single-agent-limiting-value} shows that, as $\beta\to 0$, the soft value update converges to a standard value update using the provided prior, similar to the value iterations in the MDPs~\cite{neuro-dynamic-programming}. 
At the same time, as the inverse temperature approaches zero, Lemma~\ref{lemma:single-agent-limiting-policy} also shows that the soft-optimal policy converges to the given prior regardless of the current soft-optimal Q-function estimate.

\subsection{Two-Agent Soft-Q Learning}

In this section,
we extend the previous lemmas to the two-agent case game-theoretic setting.
These results will serve an important role in showing the convergence of the SNQ2 algorithm in competitive MDP (cMDP) settings.

Consider the two-agent zero-sum case, where the policy-dependent soft value is defined in a similar manner as in~\eqref{eqn-appd:derivation-end-1}, as follows
\begin{align*}
    \V_\KL^{\pi\pl,\pi\op}(s) 
    &= \mathbb{E}_{s_0=s}^{\pi\pl,\pi\op} \Bigg[ \sum_{t=0}^{\infty} \gamma^t \bigg(\R(s_t,a_t\pl,a_t\op) 
    - \frac{1}{\beta\pl} \log \frac{\pi\pl(a_t\pl|s_t)}{\rho\pl(a_t\pl|s_t)}
    - \frac{1}{\beta\op} \log \frac{\pi\op(a_t\op|s_t)}{\rho\op(a_t\op|s_t)} \bigg)
    \Bigg]
    \\
    &=  \sum_{a\pl} \sum_{a\op} \pi\pl(a\pl|s) \pi\op(a\op|s) \Bigg( \R(s,a\pl,a\op)+\gamma \sum_{s'\in\S} \T_{s s'}(a\pl,a\op) \V_\KL^{\pi\pl,\pi\op}(s')\\
    &\quad \quad \quad \quad \quad \quad \quad \quad \quad \quad \quad 
     \quad \quad \quad \quad \quad \quad \quad \quad \quad \quad 
    -\frac{1}{\beta\pl} \log \frac{\pi\pl(a\pl|s)}{\rho\pl(a\pl|s)}
    -\frac{1}{\beta\op} \log \frac{\pi\op(a\op|s)}{\rho\op(a\op|s)} \Bigg).
\end{align*}

Define the $\left(\pi\pl,\pi\op \right)$-policy-dependent Q-value in a similar manner as in (\ref{eqn299:pt}) 
\begin{equation} \label{eq:QKL2agent}
    \Q_{\KL}^{\pi\pl,\pi\op}(s,a\pl,a\op) = \R(s,a\pl,a\op) 
    + \gamma\sum_{s'\in \S}\T_{ss'}(a\pl,a\op)\V^{\pi\pl,\pi\op}_\KL(s').
\end{equation}

Then, we have
\begin{align}
        &\V_\KL^{\pi\pl,\pi\op}(s) \\
        &= \sum_{a\pl} \Bigg[\pi\pl(a\pl|s) \Bigg\{\sum_{a\op} \left[  \pi\op(a\op|s)\bigg(\Q_{\KL}^{\pi\pl,\pi\op}(s,a\pl,a\op) 
        -\frac{1}{\beta\op} \log \frac{\pi\op(a\op|s)}{\rho\op(a\op|s)}\bigg)\right] 
        -
        \frac{1}{\beta\pl} \log \frac{\pi\pl(a\pl|s)}{\rho\pl(a\pl|s)}\Bigg\}\Bigg]. \nonumber
\end{align}

We consider the case where the Player is the maximizer and the Opponent is the minimizer. 
Then, the definition of the soft-optimal value for the two-agent zero-sum game is
\begin{align}
    &\V_\KL^\star(s) 
    = \max_{\pi\pl} \halfspace \min_{\pi\op} \V^{\pi\pl,\pi\op}_\KL(s)\label{eqn:soft-min-max}\\
    &= \max_{\pi\pl} \sum_{a\pl} \Bigg[\pi\pl(a\pl|s) \Bigg\{
    \underbrace{\min_{\pi\op} \Bigg[ \sum_{a\op}  \pi\op(a\op|s)\Bigg(\Q_\KL^*(s,a\pl,a\op)
    -\frac{1}{\beta\op} \log \frac{\pi\op(a\op|s)}{\rho\op(a\op|s)}\Bigg) \Bigg]}_{\text{Minimization problem of Opponent}} 
    -\frac{1}{\beta\pl} \log \frac{\pi\pl(a\pl|s)}{\rho\pl(a\pl|s)}\Bigg\}\Bigg] \nonumber,
\end{align}
where the soft-optimal Q-value $\Q^\star_\KL$ is defined as
\begin{equation*}
    \Q_\KL^\star(s,a\pl,a\op)  = \R(s,a\pl,a\op) 
    + \gamma\sum_{s'\in \S}\T_{ss'}(a\pl,a\op)\V^\star_\KL(s').
\end{equation*}

\begin{theorem} \label{thm:unique-soft-optimal-value}
    The soft-optimal value in~(\ref{eqn:soft-min-max}) is unique. 
    Specifically, for all states $s \in \S$,
    \begin{equation*}
        \max_{\pi\pl} \halfspace \min_{\pi\op} \V^{\pi\pl,\pi\op}_\KL(s) =
        \min_{\pi\op}\halfspace \max_{\pi\pl}  \V^{\pi\pl,\pi\op}_\KL(s).
    \end{equation*}
\end{theorem}

\begin{proof}
    One can easily check that $\V_\KL^{\pi\pl,\pi\op}$ is concave in $\pi\pl$ and convex in $\pi\op$. 
    Then, via the Minimax Theorem~\cite{vonNeumann:2007}, the soft-optimal value is unique.
\end{proof}

Similarly to the analysis for the single-agent case, consider a value function $\halfspace \U \in \Re^{|\S|}$, 
and define the update operator~$\F:\Re^{|\S|} \to \Re^{|\S|}$
as
\begin{equation}\label{eqn-appd:two-agent-bellman}
\begin{aligned}
    \big(\mathcal{F}\mathcal{U}\big)(s) 
    = \max_{\bm{\pi}\pl(s)}~ \min_{\bm{\pi}\op(s)} \Big( \R(s,\bm{\pi}\pl(s),\bm{\pi}\op(s)) + \gamma &\sum_{s'\in \S} \T_{s s'}\big(\bm{\pi}\pl(s),\bm{\pi}\op(s)\big)  \mathcal{U}(s')\\
    & -\frac{1}{\beta\pl} \mathrm{KL}(\bm{\pi}\pl(s)||\bm{\rho}\pl(s))
    -\frac{1}{\beta\op} \mathrm{KL}(\bm{\pi}\op(s)||\bm{\rho}\op(s)) \Big).
\end{aligned}
\end{equation}
It has been shown in~\cite{Balance-Two-Player-Soft-Q:2018} that the operator $\F$ is a sup-norm contraction mapping. 
As a result, the soft-optimal value function is the fixed point of $\F$. 
That is, $\V_\KL^\star = \F \halfspace \V_\KL^\star$.
One can then iterate $\F$ to find the soft-optimal value function.
Let $\V_{\KL,t}$ denote the estimate of the soft-optimal value-function at iteration $t$.
That is,
\begin{equation}
    \V_{\KL,t} = \F \V_{\KL,t-1}, \quad t=0,1,2,\ldots.
\end{equation}
Due to the contraction property of $\F$, we have
\begin{equation*}
    \V_\KL^\star = \lim_{t\to\infty} \V_{\KL,t} \halfspace.
\end{equation*}
The estimate of the soft-optimal Q-function $\Q_{\KL,t}$ is related to the soft-optimal value estimate $\V_{\KL,t}$ via 
\begin{equation*}
    \Q_{\KL,t}(s,a\pl,a\op) = \R(s,a\pl,a\op) + \gamma \sum_{s'\in \S} \T_{s s'}(a\pl,a\op)  \V_{\KL,t}(s').
\end{equation*}

We now derive the closed-form formula of the operator $\F$.
Let $\V_{\KL,t}$ and $\Q_{\KL,t}$ denote the estimates at iteration $t$.
The optimization problem in \eqref{eqn:soft-min-max} then becomes,
\begin{align}
    &\V_{\KL,t+1}(s) = \left(\F \V_{\KL,t}\right)(s) \label{appdx-eqn:soft-min-max-t} \\
    &= \max_{\pi\pl} \sum_{a\pl} \Bigg[\pi\pl(a\pl|s) \Bigg\{
    \underbrace{\min_{\pi\op} \Bigg[ \sum_{a\op}  \pi\op(a\op|s)\Bigg(\Q_{\KL,t}(s,a\pl,a\op)
    -\frac{1}{\beta\op} \log \frac{\pi\op(a\op|s)}{\rho\op(a\op|s)}\Bigg) \Bigg]}_{\text{Minimization problem of Opponent}} 
    -\frac{1}{\beta\pl} \log \frac{\pi\pl(a\pl|s)}{\rho\pl(a\pl|s)}\Bigg\}\Bigg] \nonumber.
\end{align}

We may then 
define the result of the minimization conducted by the Opponent in~(\ref{appdx-eqn:soft-min-max-t}) as 
\begin{align} \label{eqn:Q^*_PL}
    \Q_{\KL,t}\pl(s,a\pl) &= \min_{\pi\op}  \sum_{a\op \in \A\op}  \pi\op(a\op|s)\Bigg[\Q_{\KL,t}(s,a\pl,a\op) 
    -\frac{1}{\beta\op} \log \frac{\pi\op(a\op|s)}{\rho\op(a\op|s)}\Bigg]  .
\end{align}
Observe that~(\ref{eqn:Q^*_PL}) has the same structure as the optimization problem presented in~(\ref{eqn-appd:min-soft-optimal-problem}). 
Using Corollary~\ref{cor:min-optimal-policy-value}, we therefore have
\begin{align}\label{eqn:pl-marg}
    \Q_{\KL,t}\pl(s,a\pl) = \frac{1}{\beta\op} \log \sum_{a\op \in \A\op} \rho\op(s|a\op) \exp{\beta\op \Q_{\KL,t}(s,a\pl,a\op)}.
\end{align}
Combining~(\ref{eqn:soft-min-max}) and~(\ref{eqn:pl-marg}), we arrive at
\begin{equation}\label{eqn-appd:V_KL^*_player}
    \V_{\KL,t+1}(s) = \max_{\pi\pl} \sum_{a\pl \in \A\pl} \pi\pl(a\pl|s) 
    \Bigg[ \Q_{\KL,t}\pl(s,a\pl)-\frac{1}{\beta\pl} \log \frac{\pi\pl(a\pl|s)}{\rho\pl(a\pl|s)} \Bigg].
\end{equation}
Notice that~(\ref{eqn-appd:V_KL^*_player}) has the same structure as in~(\ref{eqn-appd:v_KL_t}).
By Lemma~\ref{lemma:soft-optimal-policy}, we have that the solution for the policy of the Player that maximizes \eqref{eqn-appd:V_KL^*_player} is given by
\begin{equation}\label{eqn:pl-soft-policy}
    \pi_{\KL,t}\pl(a\pl|s) = \frac{\rho\pl(a\pl|s)\exp{\beta\pl \Q_{\KL,t}\pl(s,a\pl)}}
    {\sum_{a^{\text{pl}'}\in \A\pl}\rho(a\pl|s)\exp{\beta\pl \Q_{\KL,t}\pl(s,a^{\text{pl}'})}}.
\end{equation}

Applying Corollary~\ref{cor:max-soft-optimal-value} to~(\ref{eqn-appd:V_KL^*_player}), we obtain the updated estimate of the soft-optimal value as
\begin{equation}\label{eqn-appd:soft-optimal-value-1}
    \V_{\KL,t+1}(s) = \frac{1}{\beta\pl} \log \sum_{a\pl} \rho \pl(a\pl|s)\exp{\beta\pl \Q_{\KL,t}\pl(s,a\pl)}.
\end{equation}
Since the soft-optimal value is unique, we can change the order of the $\max$ and $\min$ operators in~(\ref{eqn:soft-min-max}).
We would then have a $\min$-$\max$ optimization problem instead. 
The soft-optimal policy at some iteration $t$ of the Opponent can be then derived in a similar manner (see~\cite{Balance-Two-Player-Soft-Q:2018}).
Namely, the maximization conducted by the Player results in the following marginalization
\begin{align*}
    \Q_{\KL,t}\op(s,a\op) = \frac{1}{\beta\pl} \log \sum_{a\pl \in \A\pl} \rho\pl(s|a\pl) \exp{\beta\pl \Q_{\KL,t}(s,a\pl,a\op)},
\end{align*}
and the soft-optimal policy for the Opponent is 
\begin{equation}\label{eqn:op-soft-policy}
    \pi_{\KL,t}\op(a\op|s) = \frac{\rho\op(a\op|s)\exp{\beta\op \Q_{\KL,t}\op(s,a\op)}}
    {\sum_{a^{\text{op}'}\in \A\op}\rho(a\op|s)\exp{\beta\op \Q_{\KL,t}\op(s,a^{\text{op}'})}}.
\end{equation}
The corresponding expression for the soft-optimal value is given by
\begin{equation}\label{eqn-appd:soft-optimal-value-2}
    \V_{\KL,t+1}(s) = \frac{1}{\beta\op} \log \sum_{a\op \in\A\op} \rho \op(a\op|s)\exp{\beta\op \Q_{\KL,t}\op(s,a\op)}.
\end{equation}

We provide the following lemma regarding the limiting behavior of the soft-optimal value and soft-optimal policies as the inverse temperatures of the Player and the Opponent both approach zero simultaneously.
\begin{lemma}\label{lemma:zero-inverse-temp-value-policies}
As the inverse temperatures $\beta\pl$ and $\beta\op$ approach 
zero,
we have the following limiting relationship regarding the soft-optimal value and the policies
\begin{align}
    &\lim_{(\beta\pl,\beta\op)\to (0,0)} \V_{\KL,t+1}(s) = \sum_{a\pl} \sum_{a\op} \rho\pl(s,a\pl)\rho\op(s,a\op) \Q_{\KL,t}(s,a\pl,a\op), \label{appd-eqn:limiting-value}\\
    &\lim_{(\beta\pl,\beta\op)\to (0,0)}  \pi_{\KL,t}\pl (a\pl|s) = \rho\pl(a\pl|s),
    \label{appd-eqn:limiting-soft-policy-pl}\\ 
    &\lim_{(\beta\pl,\beta\op)\to (0,0)}  \pi_{\KL,t}\op(a\op|s) = \rho\op(a\op|s).\label{appd-eqn:limiting-soft-policy-op}
\end{align} 
\end{lemma}

\begin{proof}
We first show that $\V_{\KL,t}$ is continuous with respect to $\beta\pl$ and $\beta\op$.
Consider the following log-sum-exp function, where $\bm{x} = (x_1,x_2,\ldots, x_n)$
\begin{equation} \label{eq:logexp}
    \phi(\beta, \bm{x}) = \frac{1}{\beta} \log \sum_i \lambda_i \mathrm{exp}\, (\beta x_i) ,
\end{equation}
where $\lambda_i>0$ and $\sum_i \lambda_i = 1$.
One can easily show that (\ref{eq:logexp}) is a continuous function with respect to $\beta$ and $\bm{x}$.


The function $\V_{\KL,t}$ can then be written as follows
\begin{equation}\label{eqn:composition}
    \V_{\KL,t}= \phi\Big(\beta\pl, \Big[\phi\big(\beta\op, \bm{Q}_{\KL}(1,\cdot)\big),\ldots \phi\big(\beta\op, \bm{Q}_{\KL}(|\A\pl|,\cdot)\big)\Big]\Big),
\end{equation}
where $\bm{Q}_{\KL}(a\pl,\cdot)$ is a row vector.
Since $\phi$ is continuous, the composition in~\eqref{eqn:composition} is also continuous with respect to $\beta\pl$ and $\beta\op$. 
We can show the continuity of \eqref{appd-eqn:limiting-soft-policy-pl} and \eqref{appd-eqn:limiting-soft-policy-op} through a similar argument.

Now, with the continuity of $\V_{\KL,t}$ established, we can prescribe a specific 
trajectory%
\footnote{%
For the SNQ2 algorithm, the trajectory of $(\beta\pl,\beta\op)\to (0,0)$ needs to be restricted to the open fourth quadrant, i.e. $\beta\pl>0$ and $\beta\op<0$, so that the game has the max-min structure. 
However, since the functions $\V_{\KL,t+1}, \pi_{\KL,t}\pl$ and $\pi_{\KL,t}\op$ in Lemma~\ref{lemma:zero-inverse-temp-value-policies} are continuous with respect to $(\beta\pl,\beta\op)$,
the particular trajectory followed is inconsequential, as it results in the same limit.}
along which $(\beta\pl,\beta\op)$ approaches $(0,0)$. 
To mimic the manner the SNQ2 algorithm updates the values of $\beta\pl$ and $\beta\op$, 
we choose the following path parameterized by $\beta>0$:
\begin{equation}\label{appd-eqn:beta-traj}
    \beta\pl = \beta, \quad \beta\op = - \kappa \beta, \quad~\text{ such that }~ \beta \to 0^+ ~\text{ and }~ \kappa >0.
\end{equation}
To show \eqref{appd-eqn:limiting-value}, notice that
\begin{align*}
    \lim_{\beta \to 0^+}&\V_{\KL,t+1}(s) = \lim_{\beta \to 0^+} \frac{1}{\beta} \log \sum_{a\pl} \rho \pl(a\pl|s)\exp{\frac{\beta}{-\kappa \beta} \log \sum_{a\pl} \rho\pl(s|a\pl) \exp{-k \beta \Q_{\KL,t}(s,a\pl,a\op)}}\\
    &= \lim_{\beta \to 0^+} \frac{1}{\beta} \log \sum_{a\pl} \rho \pl(a\pl|s)
    \left(\sum_{a\pl} \rho\pl(s|a\pl) \exp{-\kappa \beta \Q_{\KL,t}(s,a\pl,a\op)}\right)^{-{1}/{\kappa}} \\
    &= \lim_{\beta \to 0^+} \frac{ \frac{\partial}{\partial \beta}\log \sum_{a\pl} \rho \pl(a\pl|s)
    \left(\sum_{a\pl} \rho\pl(s|a\pl) \exp{-\kappa \beta \Q_{\KL,t}(s,a\pl,a\op)}\right)^{-{1}/{\kappa}}}{ \frac{\partial}{\partial \beta} \beta}\\
    &= \sum_{a\pl} \sum_{a\op} \rho\pl(s,a\pl)\rho\op(s,a\op) \Q_{\KL,t}(s,a\pl,a\op).
\end{align*}
Similarly, for the numerator of the Player's soft-optimal policy in \eqref{eqn:pl-soft-policy}, we have
\begin{align}
    \lim_{\beta \to 0^+} &\rho\pl(a\pl|s)\exp{\beta\pl \Q_{\KL,t}\pl(s,a\pl)} \\
    &= \lim_{\beta \to 0^+} \rho\pl(a\pl|s)\exp{ \frac{\beta\pl}{\beta\op} \log \sum_{a\op} \rho\op(s|a\op) \exp{\beta\op \Q_{\KL,t}(s,a\pl,a\op)}}\\
    &= \rho\pl(a\pl|s) \left(\sum_{a\op} \rho\op(s|a\op) \lim_{\beta \to 0^+}  \exp{-\kappa \beta \Q_{\KL,t}(s,a\pl,a\op)}\right)^{-{1}/{\kappa}}\\
    &= \rho\pl(a\pl|s) \left(\sum_{a\op} \rho\op(s|a\op) \cdot 1\right)
    =\rho\pl(a\pl|s)
\end{align}
Next, for the soft-optimal policy $\pi\pl_{\KL,t}$, we have
\begin{align*}
    \lim_{\beta \to 0^+} \pi_{\KL,t}\pl(a\pl|s) &= \frac{\lim_{\beta \to 0}\rho\pl(a\pl|s)\exp{\beta\pl \Q_{\KL,t}\pl(s,a\pl)}}
    {\sum_{a^{\text{pl}'}}\lim_{\beta \to 0^+}\rho(a\pl|s)\exp{\beta\pl \Q_{\KL,t}\pl(s,a^{\text{pl}'})}}\\
    &= \frac{\rho\pl(a\pl|s)}{\sum_{a^{\text{pl}'}}\rho(a^{\text{pl}'}|s)} = \rho\pl(a\pl|s).
\end{align*}
One can show \eqref{appd-eqn:limiting-soft-policy-op} in a similar manner.
\end{proof}

\section{Sequential Policy Approximation}

In this section, we present the SNQ2 algorithm in a cMDP setting. 
That is, we assume that the environment 
is known to the agents, and thus no stochastic approximation is used. 
We show the convergence of the SNQ2 algorithm in a cMDP setting
in the next section.
We first define the metric spaces used for the proof of convergence of the SNQ2 algorithm.

\subsection{Metric Spaces}

\paragraph{Metric Space of $\Q$-functions.}  
As our algorithm operates with two separate $Q$-functions, we denote the standard $Q$-function as $\Q$ and the soft $Q$-function as $\Q_\KL$ to differentiate between the two, even though both live in the same space $Q \subseteq \Re^{|S|\times |\A\pl| \times |\A\op|}$.
We equip the vector space $Q$ with the distance metric $\D$ 
induced by the sup-norm.
That is, for $\Q,\Q' \in Q$,
\begin{equation*}
    \D(\Q, \Q') = \norm*{\Q-\Q'}_\infty = \max_{s,a\pl,a\op} \abs*{\Q(s,a\pl,a\op)-\Q'(s,a\pl,a\op)}. 
\end{equation*}
Note that $(Q,\D)$ is a complete metric space.

\paragraph{Metric Space of concatenated $Q$-functions.}
We first define the concatenated $Q$-function as
\begin{equation*}
    \bar{\Q} = \left[\begin{array}{c}
         \Q_{~}\\
         \Q_\KL 
    \end{array}\right].
\end{equation*}
We then use $\bar{Q} \subseteq \Re^{2\times|\S| \times |\A\pl| \times |\A|\op}$ as the vector space for the concatenated $Q$-functions.
We equip the vector space $\bar{Q}$ with  the distance metric $\bar{\D}$ defined as follows:
\begin{equation*}
    \bar{\D}(\bar{\Q},\bar{\Q}') = \D(\Q,\Q') + \D(\Q_\KL, \Q_\KL').
\end{equation*}
One can easily verify the identity of indiscernibles, the  symmetry, and 
subadditivity properties of $\bar{\D}$. 

\paragraph{Metric Space of Policy Pairs.}  
Similar to the $Q$-functions, the standard policy pair $\pi$ and the soft policy pair $\pi_\KL$ both live in the same metric space $\Pi = \big(\Delta_{|\A\pl|}\big)^{|S|} \times \big(\Delta_{|\A\op|}\big)^{|S|}$.
The policy $\bm{\pi}\pl(s)$ of the Player at state $s$ lives on the standard simplex $\Delta_{|\A\pl|}$. 
With a total of $|\S|$ states, the policy $\pi\pl$ is an element of $\big(\Delta_{|\A\pl|}\big)^{|S|}$.
A  similar argument holds for $\pi\op$.
We define a metric $\W$ based on the total variation between two policy pairs~\cite{neurips-mean-field}
\begin{equation}
    \W(\pi, \pi') = \max_s \{\dTV {\bm{\pi}_1\pl(s), \bm{\pi}_2\pl(s)} \} 
    + \max_s \{\dTV{\bm{\pi}_1\op(s), \bm{\pi}\op_2(s)}\}.
\end{equation}
Since the action spaces $\A\pl$ and $\A\op$ are discrete, the total variation between the two policies  $\bm{\pi}_1(s),\bm{\pi}_2(s) \in \Delta_{|\A|}$ at state $s$ can be easily computed as~\cite{stochastic-processes:ross} 
\begin{equation*}
    \dTV{\bm{\pi}_1(s),\bm{\pi}_2(s)} = \frac{1}{2}\sum_{a\in\A} \Big \vert \pi_1(a|s)-\pi_2(a|s)\Big \vert.
\end{equation*}


\subsection{Basic Operators}

Before presenting the SNQ2 algorithm, we summarize the four operators that are used in the algorithm.
Note that the superscript $\beta$ on $\Gamma^\beta_2$ and $\Gamma^\beta_\KL$ is a scalar parameter used to parameterize the trajectory of $(\beta\pl,\beta\op)$ approaching $(0,0)$, via $\beta = \beta\pl = - \beta\op/\kappa$ with $\kappa > 0$ by letting $\beta \to 0^+$.
The setup in~\eqref{appd-eqn:beta-traj} serves an example.

\paragraph{Standard Q-update Operator:} $\Gamma_1:Q \times \Pi \to Q$.
\\
    Given a $Q$-function, and a policy pair $\pi = (\pi\pl,\pi\op)$, 
    the standard Q-update Operator $\Gamma_1$ updates the $Q$-function via:
    \begin{equation}
        \Big[\Gamma_1(\Q,\pi)\Big](s,a\pl,a\op) = \R(s,a\pl,a\op) +\gamma \sum_{s'} \T_{ss'}(a,\pl,a\op) \left[\bm{\pi\pl}(s')\trps \bm{\Q}(s') \bm{\pi\op}(s')\right].
    \end{equation}
    
\paragraph{Soft Q-update Operator:}
$\Gamma^\beta_2: Q \times \Pi \to Q$.
\\
    Given a soft $Q$-function $\Q_\KL$, and a reference policy pair $\rho = (\rho\pl,\rho\op)$, 
    the soft update operator performs the following update:
    \begin{equation}
        \begin{aligned}
        &\Big[\Gamma^\beta_2(\Q_\KL,\rho)\Big](s,a\pl,a\op) \\
        & \qquad \qquad \qquad = \R(s,a\pl,a\op) +\frac{\gamma}{\beta\pl} \sum_{s'} \T_{ss'}(a,\pl,a\op) \log \sum_{a\pl} \rho \pl(a\pl|s)\exp{\beta\pl \Q_{\KL}\pl(s,a\pl)},
    \end{aligned}
    \end{equation}
    where $ \Q_{\KL}\pl$ is obtained, via
    marginalization, from
    \begin{equation*}
        \Q_{\KL}\pl(s,a\pl) = \frac{1}{\beta\op} \log \sum_{a\op \in \A\op} \rho\op(s|a\op) \exp{\beta\op \Q_{\KL}(s,a\pl,a\op)}.
    \end{equation*}

\paragraph{Soft Q-update Operator:} $\Gamma_\Nash: Q \to \Pi$.
\\
    Given a $Q$-function, the operator $\Gamma_\Nash$ finds the 
    Nash policies corresponding to this $Q$ function by solving 
    the following two linear programs~\cite{cMDP:2012} 
    \begin{equation*}
    \begin{alignedat}{4}
        &\quad \quad \max \quad                 && v,                                                        
       &&\quad \quad \min \quad                 && u,\\ 
        & \text{subject to} \quad  && v\mathds{1}\trps - \bm{\pi}\pl(s) \trps \bm{\Q}(s) \leq 0, \quad \quad \quad \quad 
       && \text{subject to} \quad  && u\mathds{1} - \bm{\Q}(s) \bm{\pi}\op(s) \geq 0,\\
        &  \quad                    && \mathds{1}\trps \bm{\pi}\pl(s) = 1, 
        ~~ \bm{\pi}\pl(s) \geq 0,
       &&  \quad                    && \mathds{1}\trps \bm{\pi}\op(s) = 1, 
       ~~ \bm{\pi}\op(s) \geq 0.
    \end{alignedat}
    \end{equation*}
    That is,
      \begin{equation}
        \pi_\Nash = (\pi\pl_\Nash,\pi\op_\Nash) = \Gamma_\Nash \halfspace \Q.
    \end{equation}    
    

\paragraph{Soft-Policy Generating Operator:} $\Gamma^\beta_\KL: \Q \times \Pi \to \Pi$.
\\
Given a $Q$-function $\Q_\KL$, and a reference policy pair $\rho = (\rho\pl,\rho\op)$, 
one can compute the corresponding soft-optimal policies for the Player and the Opponent per \eqref{eqn:pl-soft-policy} and \eqref{eqn:op-soft-policy}, respectively.
That is, 
\begin{align*}
    \pi_{\KL}\pl(a\pl|s) &= \frac{\rho\pl(a\pl|s)\exp{\beta\pl \Q_{\KL}\pl(s,a\pl)}}
    {\sum_{a^{\text{pl}'}\in \A\pl}\rho(a\pl|s)\exp{\beta\pl \Q_{\KL}\pl(s,a^{\text{pl}'})}}\\
    \pi_{\KL}\op(a\op|s) &= \frac{\rho\op(a\op|s)\exp{\beta\op \Q_{\KL}\op(s,a\op)}}
    {\sum_{a^{\text{op}'}\in \A\op}\rho(a\op|s)\exp{\beta\op \Q_{\KL}\op(s,a^{\text{op}'})}}
\end{align*}
We then can define the soft-policy generating operator $\Gamma^\beta_\KL$ as
\begin{equation}
    \pi_\KL = (\pi_\KL\pl,\pi_\KL\op) = \Gamma_\KL^\beta(\Q_\KL,\rho).
\end{equation}

\subsection{The SNQ2 Algorithm in cMDP Setting}

Algorithm~\ref{alg:cMDP-SNQ2-appdx} presents the non-learning baseline version of the SNQ2 algorithm.
We present the convergence proof of the baseline SNQ2 in the next section. 

\begin{algorithm}[h] 
\SetAlgoLined
\caption{Baseline SNQ2 in cMDP Setting}
\footnotesize
\label{alg:cMDP-SNQ2-appdx}
\lstset{numbers=left, numberstyle=\tiny, stepnumber=1, numbersep=5pt}
\footnotesize
\textbf{Inputs:} Game tuple $\G =\langle \mathcal{S}, \mathcal{A}\pl, \mathcal{A}\op, \mathcal{T},\mathcal{R},\gamma \rangle$\;
Set $\Q(s,a\pl,a\op) = \Q_{\KL}(s,a\pl,a\op) = 0$ for all states $s$ and actions $a\pl, a\op$\;
Set $\beta\pl$ and $\beta \op$ to some large values\; \label{Q2-Alg-init-end-appdx}
\While{$\Q$ and $\Q_\KL$ \text{ not converged}}{
    Update Nash policies: $\pi_\Nash \leftarrow \Gamma_\Nash(\Q)$; \\
    Update soft optimal policies: $\pi_\KL \leftarrow \Gamma^\beta_\KL(\Q_\KL,\pi_\Nash)$; \\
    Update Q-function: $\Q\_ \leftarrow \Gamma_1(\Q, \pi_\KL)$; \\
    Update soft Q-function: $\Q_\KL\_ \leftarrow \Gamma^\beta_2(\Q_\KL, \pi_\Nash)$; \\
    $\Q \leftarrow \Q\_$, $\Q_\KL \leftarrow \Q_\KL \_$; \\
    Reduce inverse temperature $\beta$
}
\Return $\pi_\Nash$ and $\Q(s,a\pl,a\op)$.
\end{algorithm}

This baseline version of the SNQ2 serves as a demonstration of the sequential policy approximation behind the SNQ2 algorithm. 
It does not alleviate the computational burden of computing Nash policies, as the Nash policy is updated at every iteration.
However, it is easier to show the convergence of the baseline SNQ2 algorithm to the Nash Q-function.
The convergence of the general SNQ2 algorithm then follows readily using a simple argument (see Section~\ref{subsec:conv-standard-SNQ2} and, in particular, Theorem~\ref{thm:conv-standard-SNQ2})

In Algorithm~\ref{alg:cMDP-SNQ2-full}, we present the standard SNQ2 algorithm in the cMDP setting. 
Different from the baseline version, the standard SNQ2 updates the prior for the soft optimal policy every $M$
iterations of Q-function updates. 

The algorithm consists of two iteration loops: the inner loop (from line 7 to line 10) updates the Q-functions for a prescribed number $M$; the outer loop (from line 5 to line 12) computes the Nash policies fed into $\Gamma_2^\beta$ and $\Gamma^\beta_\KL$, and it also reduce the inverse temperature. 

\begin{algorithm}[h]
\SetAlgoLined
\caption{SNQ2 in cMDP Setting}
\footnotesize
\label{alg:cMDP-SNQ2-full}
\lstset{numbers=left, numberstyle=\tiny, stepnumber=1, numbersep=5pt}
\footnotesize
\textbf{Inputs:} Game tuple $\G =\langle \mathcal{S}, \mathcal{A}\pl, \mathcal{A}\op, \mathcal{T},\mathcal{R},\gamma \rangle$\;
Set $\Q(s,a\pl,a\op) = \Q_{\KL}(s,a\pl,a\op) = 0$ for all states $s$ and actions $a\pl, a\op$\;
Set $\beta\pl$ and $\beta \op$ to some large values\;
\While{$\Q$ and $\Q_\KL$ \text{ not converged}}{
    Update Nash policies: $\pi_\Nash \leftarrow \Gamma_\Nash(\Q)$; \\
    \For{InnerIteration = 1 ~to~ M}{
    Update soft optimal policies: $\pi_\KL \leftarrow \Gamma^\beta_\KL(\Q_\KL,\pi_\Nash)$; \\
    Update Q-function: $\Q\_ \leftarrow \Gamma_1(\Q, \pi_\KL)$; \\
    Update soft Q-function: $\Q_\KL\_ \leftarrow \Gamma^\beta_2(\Q_\KL, \pi_\Nash)$; \\
    $\Q \leftarrow \Q\_$, $\Q_\KL \leftarrow \Q_\KL \_$; \\
    }
    Reduce inverse temperature $\beta$
}
\Return $\pi_\Nash$ and $\Q(s,a\pl,a\op)$.
\end{algorithm}

\section{Convergence of the SNQ2 Algorithm}

In this section, some regularity assumptions on the operators are introduced for the analysis of convergence. 
The main theorem regarding the convergence of sequentially applying a family of contraction mappings is then presented in Theorem~\ref{thm:sequential-application-contraction}.
Based on that theorem, we first show the convergence of the baseline SNQ2 and then the convergence of the (standard) SNQ2.

\subsection{Convergence of Sequentially Applying a Family of Contraction Mappings}

The SNQ2 algorithm differs from standard iterative algorithms that utilize contraction mapping arguments, in the sense that SNQ2 utilizes a different contracting operator to update the concatenated Q-function at each iteration, as $\beta$ decreases. 
Thus, we need the following theorem regarding the convergence of iteratively
applying a family of contraction mappings.

\begin{theorem} \label{thm:sequential-application-contraction}
    Let $(\X,\rho)$ be a complete metric space, let $f_n: \X \to \X$ be a family of $\rho$-contractions, 
    such that, for all $n=1,2,\ldots$ there exists $d_n\in(0,1)$, such that
    $\rho(f_n x, f_n y) \leq d_n \halfspace \rho(x, y)$ for all $x, y \in \X$.
    Assume that $\lim_{n \to \infty} d_n = d \in (0,1)$.
    Let $x \in \X$, and let $x^{(n)} = f_n\cdots f_1 x$ be the result of sequentially applying the operators $f_1, \ldots, f_n$ to $x$.
    If the sequence of operators $\{f_n\}_{n=1}^\infty$ converges pointwise to $f$, 
    then $f$ is also a $\rho$-contraction mapping with contraction factor $d$.
    Furthermore, if $x^\star$ is the fixed point of $f$,
    then, for every $x\in \X$, $\lim_{n \to \infty} x^{(n)} = x^\star$.
\end{theorem}

\begin{proof}
We first show that the limiting operator $f$ is a contraction mapping. 
Let $x,y\in \X$.
From the pointwise converge of operators $\{f_n\}_{n=1}^\infty$
it follows that
\begin{equation*}
    \lim_{n\to \infty} f_n x = f x,  \qquad
    \lim_{n\to \infty} f_n y = f y. 
\end{equation*}
Since $\rho$ is continuous~\cite{topology}, it follows that
\begin{align*}
    \rho(fx, fy) &= \rho ( \lim_{n\to \infty} f_n x, \lim_{n\to \infty} f_n y ) \\
    &=  \lim_{n\to\infty}\rho(f_n x ,f_n y ) \\
    &\leq \lim_{n\to \infty} d_n \, \rho(x,y) \\
    & = d \, \rho(x,y),
\end{align*}


Since the limiting operator $f$ is a contraction mapping, 
its fixed point $x^\star = f x^\star$ is unique. 
Let $x\in \X$. 
We then have
\begin{align*}
    \rho(x^{(n)}, x^\star) &= \rho(f_n\cdots f_1 x, \underbrace{f \cdots f}_{n} x^\star) \\
    & \leq \rho(f_n\cdots f_1 x , f_n \underbrace{f \cdots f}_{n-1} x^\star) + 
    \rho(f_n \underbrace{f \cdots f}_{n-1} x^\star , \underbrace{f \cdots f}_{n} x^\star) \\
    &= \rho(f_n f_{n-1}\cdots f_1 x , f_n \underbrace{f \cdots f}_{n-1} x^\star) + 
    \rho(f_n x^\star, f x^\star).
\end{align*}
Since the sequence $\{f_n\}_{n=1}^\infty$ converges pointwise to $f$, we have for the second term, that
\begin{equation*}
 \lim_{n\to\infty}\rho(f_n x^\star , f x^\star) = 
 \rho( \lim_{n\to\infty} f_n x^\star , f x^\star) = 
 \rho( f x^\star , f x^\star) =
 0.  
\end{equation*}

Since $f_n$ is a $\rho$-contraction with a contraction factor $d_n \in (0,1)$, 
we have for the first term that
\begin{equation*}
    \rho(f_n f_{n-1} \cdots f_1 x , f_n \underbrace{f \cdots f}_{n-1} x^\star) \leq d_n \, \rho(f_{n-1}\cdots f_1 x , \underbrace{f \cdots f}_{n-1} x^\star) 
    = d_n \, \rho(x^{(n-1)},x^\star). 
\end{equation*}

Let $y_n = \rho(x^{(n)} , x^\star)$ and $z_{n-1} =\rho(f_n x^\star , f x^\star)$. 
From the previous series of inequalities, we have shown that
\begin{equation} \label{appd-eqn:induction-sequence}
    y_{n+1} \leq d_n \, y_{n} + z_{n},
\end{equation}
where, 
\begin{equation*}
   \lim_{n \to \infty}  z_n = 0  \qquad 
\lim_{n \to \infty}    d_n = d,
\end{equation*}
where $ d \in (0,1)$.
Using (\ref{appd-eqn:induction-sequence}), along with $y_n \ge0$ and $z_n \ge 0$ yields
\begin{equation} \label{yn:eq1}
    y_n \le \prod_{k=1}^{n-1} d_k y_1 + \sum_{k=1}^{n-1} \prod_{\ell = k+1}^{n-1} d_\ell z_k,
\end{equation}
with the convention
\begin{equation}
    \prod_{\ell = n}^{n-1} d_\ell = 1.
\end{equation}
Since $d_k \to d$ where $0 < d < 1$, it follows that there exists 
$N>0$ such that, for all $k \ge N$, $d_k < \eta < 1$.
The first term in (\ref{yn:eq1}) can then be bounded as
\begin{equation}
    \prod_{k=1}^{n-1} d_k y_1 = 
    \prod_{k=1}^{N-1} d_k 
     \prod_{k=N}^{n-1} d_k 
    y_1 
\le 
\alpha_1 \eta^{n},
\end{equation}
where $\alpha_1 = \prod_{k=1}^{N-1} d_k y_1/\eta^N$.
Let $k$ be fixed. 
The second term in (\ref{yn:eq1}) can be bounded as follows
\begin{equation}
    \prod_{\ell=k+1}^{n-1} d_\ell z_k = 
    \prod_{\ell=k+1}^{N-1} d_\ell 
     \prod_{\ell=N}^{n-1} d_\ell 
    z_k
    \le 
     \alpha_{2,k} \eta^{n},
\end{equation}
where $\alpha_{2,k} = \prod_{\ell=k+1}^{N-1} d_\ell z_k /\eta^N \leq z_k/\eta^N$.
Since $z_n \to 0$ as $n \to \infty$, it follows that there exists $\bar{z}$ such that $|z_n| \le \bar{z}$ for all $n \ge 1$.
In particular,
\begin{equation}
    \sum_{k=1}^{n-1} \prod_{\ell = k+1}^{n-1} d_\ell z_k \le (\bar{z}/\eta^N)
    \sum_{k=1}^{n-1} \eta^{n} = (\bar{z}/\eta^N) (n-1) \eta^n.
\end{equation}
Taking limits of both sides of (\ref{yn:eq1}) as $n \to \infty$ and $k \to \infty$, and using the fact that, for  $0 < x < 1$, $\lim_{n \to \infty} n x^n = 0$, yields 
\begin{equation}
    \lim_{n \to \infty}  y_n =   \lim_{n\to \infty} \rho(x^{(n)}, x^\star) = 0,
\end{equation}
and thus $\lim_{n \to \infty} x^{(n)} = x^\star$.

\end{proof}


\subsection{Convergence of the basedline SNQ2 Algorithm}

In this section, we present the convergence proof of the baseline SNQ2. 
We denote the operator for each iteration of the while loop of Algorithm~1 
as $\Gamma^\beta: \bar{\Q} \to \bar{\Q}$, such that
\begin{equation}    \label{eq:appC:first} 
    \Gamma^\beta \bar{\Q} = \Gamma^\beta 
    \left[
\begin{array}{c}
     \Q \\ ~\Q_\KL
\end{array}
    \right]
    =
    \left[
\begin{array}{c}
     \Gamma_1(\Q,\Gamma_\KL^\beta (\Q_\KL, \Gamma_\Nash \Q))  \\
     \Gamma_2^\beta(\Q_\KL, \Gamma_\Nash \Q)
\end{array}
    \right].
\end{equation}

In order to show the desired result, we need certain regularity assumptions on the operators to show that $\Gamma^\beta$ is a family of contraction mappings. 
Specifically, we assume that all the four operators are Lipschitz continuous with respect to their arguments.
In our experiments, however, the algorithm converges even when some of the regularity assumptions are violated at some iteration steps.

\begin{restatable}{assumption}{AssmpOne} \label{assmpt:Gamma1-Reg}
There exist constants\footnote{One can show, in particular, that $d_1 \leq \gamma$ and $d_2 \leq 2\gamma \R_{\max}/(1-\gamma)$.} 
$d_1, d_2 \geq 0$ such that for any admissible policies $\pi$ and $\pi'$ and any Q-functions $\Q$ and $\Q'$,
\begin{align}
    \D(\Gamma_1(\Q,\pi),\Gamma_1(\Q', \pi)) &\leq d_1 \D(\Q,\Q'), \\
    \D(\Gamma_1(\Q,\pi),\Gamma_1(\Q, \pi')) &\leq d_2 \W(\pi,\pi').
\end{align}
\end{restatable}

\begin{restatable}{assumption}{AssmpTwo} \label{assmpt:Gamma2-Reg}
For each $\beta$, there exist constants $d_3, d_4 \geq 0$ such that, for every admissible policies $\pi$ and $\pi'$ and every soft Q-functions $\Q$ and $\Q'$,
\begin{align}
    \D(\Gamma_2^\beta(\Q,\pi),\Gamma_2^\beta(\Q', \pi)) &\leq d_3 \D(\Q,\Q'), \\
    \D(\Gamma_2^\beta(\Q,\pi),\Gamma_2^\beta(\Q, \pi')) &\leq d_4 \W(\pi,\pi').
\end{align}
\end{restatable}

\begin{restatable}{assumption}{AssmpThree} \label{assmpt:GammaNash-Reg}
There exists a constant $d_5 \geq 0$ such that, for every Q-functions $\Q$ and $\Q'$,
\begin{align}
    \W(\Gamma_\Nash \Q, \Gamma_\Nash \Q') \leq d_5 \D(\Q, \Q').
\end{align}
\end{restatable}

\begin{restatable}{assumption}{AssmpFour} \label{assmpt:GammaKL-Reg}
For each $\beta$, there exist constants $d_6, d_7 \geq 0$ such that, for every admissible policies $\pi$ and $\pi'$ and every soft Q-functions $\Q_\KL$ and $\Q_\KL'$, 
\begin{align}
    \W(\Gamma_\KL^\beta(\Q, \pi), \Gamma_\KL^\beta(\Q, \pi'))
    &\leq d_6 \W(\pi,\pi'), \\
    \W(\Gamma_\KL^\beta(\Q, \pi), \Gamma_\KL^\beta(\Q', \pi))
    &\leq d_7 \D(\Q,\Q').
\end{align}
\end{restatable}

\begin{assumption} \label{assmpt:Lip-Const-relationship}
    The Lipschitz constants $d_i~i=1,\ldots,7$ of the operators $\Gamma_1, \Gamma_2^\beta, \Gamma_\Nash, \Gamma_\KL^\beta$ satisfy
    \begin{equation} \label{ddef}
       d = \max \{d_1+ d_4 d_5 + d_2 d_5 d_6, d_3+d_2 d_7\} < 1. 
    \end{equation}
\end{assumption}

In the rest of this section, we prove the following convergence theorem regarding the baseline SNQ2 algorithm in the cMDP setting.  

\begin{restatable}{theorem}{convMDP} \label{thm:convergence-baseline-SNQ2}
    Under Assumption \ref{assmpt:Lip-Const-relationship}, the baseline SNQ2 Algorithm in the cMDP setting converges to the Nash Q-function. 
\end{restatable}

The roadmap for the proof is as follows: 
We first show that $\Gamma^\beta$ is a family of contraction mappings under certain assumptions. 
We use this result to construct the limiting operator of $\Gamma^\beta$ when $\beta$ approaches zero, and show that the limiting operator corresponds to a standard Minimax update operator~\cite{cMDP:2012}.
We finally apply Theorem~\ref{thm:sequential-application-contraction} and show that the 
algorithm converges to the same Q-function as the Shapley's method (see Section~3.1 of the main text), which uses the standard Minimax update operator.

Next, we present the following theorem that provides sufficient conditions for $\Gamma^\beta$ to be a family of contraction mappings.

\begin{theorem}\label{thm:contraction_Gamma_beta}
    Under Assumption~\ref{assmpt:Lip-Const-relationship}, the operator $\Gamma^\beta$ is a family of $\bar{\D}$-contraction mappings with contraction factor $d$, where $d$ is given in (\ref{ddef}).
\end{theorem}

\begin{proof}
Let $\beta$ be fixed.    
For any two concatenated Q-functions $\bar{\Q}, \bar{\Q}'$, one has
    \begin{equation}\label{appdx-eqn:split-total-distance}
        \begin{aligned}
            \bar{\D}&(\Gamma^\beta \bar{\Q}, \Gamma^\beta \bar{\Q}') \\
            &=  \D(\Gamma_1(\Q,\Gamma_\KL^\beta(\Q_\KL, \Gamma_\Nash \Q)),\Gamma_1(\Q',\Gamma_\KL^\beta(\Q_\KL', \Gamma_\Nash \Q'))) +
            \D(\Gamma_2(\Q_\KL, \Gamma_\Nash \Q),\Gamma_2(\Q_\KL', \Gamma_\Nash \Q')).
        \end{aligned}
    \end{equation}

For the first term in \eqref{appdx-eqn:split-total-distance}, 
we have
\begin{equation}
    \begin{aligned}
        \D(\Gamma_1(\Q,\Gamma_\KL^\beta(\Q_\KL, \Gamma_\Nash \Q)),& \Gamma_1(\Q',\Gamma_\KL^\beta (\Q_\KL', \Gamma_\Nash \Q')))  \\
        & \leq \D(\Gamma_1(\Q,\Gamma_\KL^\beta(\Q_\KL, \Gamma_\Nash \Q)),\Gamma_1(\Q,\Gamma_\KL^\beta(\Q_\KL', \Gamma_\Nash \Q))) \\
        & ~~ + \D(\Gamma_1(\Q,\Gamma_\KL^\beta(\Q_\KL', \Gamma_\Nash \Q)),\Gamma_1(\Q,\Gamma_\KL^\beta(\Q_\KL', \Gamma_\Nash \Q'))) \\
        & ~~ + \D(\Gamma_1(\Q,\Gamma_\KL^\beta(\Q_\KL', \Gamma_\Nash \Q')),\Gamma_1(\Q',\Gamma_\KL^\beta(\Q_\KL', \Gamma_\Nash \Q'))) \\
        & \leq d_2 d_7 \D(\Q_\KL, \Q_\KL') + d_2 d_5 d_6 \D(\Q, \Q') +  d_1 \D(\Q, \Q').
    \end{aligned}
\end{equation}

For the second term in \eqref{appdx-eqn:split-total-distance},
we have
\begin{equation}
    \begin{aligned}
        \D(\Gamma_2^\beta(\Q_\KL, \Gamma_\Nash \Q),& \Gamma_2^\beta(\Q_\KL', \Gamma_\Nash \Q'))  \\
        & \leq \D(\Gamma_2^\beta(\Q_\KL, \Gamma_\Nash \Q),\Gamma_2^\beta(\Q_\KL, \Gamma_\Nash \Q'))
        + \D(\Gamma_2^\beta(\Q_\KL, \Gamma_\Nash \Q'),\Gamma_2^\beta(\Q_\KL', \Gamma_\Nash \Q')) \\
        & \leq d_4 d_5 \D(\Q,\Q') + d_3 \D(\Q_\KL,\Q_\KL').
    \end{aligned}
\end{equation}

Using the definition \eqref{ddef}, 
we have
\begin{equation}
\begin{aligned}
    \D(\Gamma^\beta \bar{\Q}, \Gamma^\beta \bar{\Q}') &\leq (d_1 + d_4 d_5 + d_2 d_5 d_6) \D(\Q,\Q') + (d_3 + d_2 d_7) \D(\Q_\KL,\Q_\KL') \\
    & \leq d \D(\Q,\Q') + d \D(\Q_\KL,\Q_\KL') = d \halfspace \D(\bar{\Q}, \bar{\Q'}).
\end{aligned}
\end{equation}
Since, by assumption $d<1$, it follows that $\Gamma^\beta$ is a family of contraction mappings.
\end{proof}

We now construct the limiting operator $\Gamma^0 = \lim_{\beta \to 0^+} \Gamma^\beta$, where $\beta$ is the scalar parameter that parameterizes the trajectory of $(\beta\pl,\beta\op)$ approaching (0,0); see, for instance~\eqref{appd-eqn:beta-traj}.
In the sequel, we use $\beta_t$ to denote the inverse temperatures used at iteration $t$, such that $\lim_{t\to \infty}\beta_t=0$ with $\beta_t > 0$.

\begin{lemma}\label{lemma:limiting-Gamma_1}
Consider a sequence of inverse temperature parameterized by $\{\beta_t\}_{t=1}^\infty$ with $\beta_t > 0$, such that $\lim_{t \to \infty} \beta_t =0$.
Then, for every Q-function $\Q\in Q$ and admissible policy pair $\pi \in \Pi$,
    \begin{equation*}
        \lim_{t \to \infty} \Gamma_\KL^{\beta_t} (\Q,\pi) = \lim_{\beta\to 0^+} \Gamma_\KL^{\beta} (\Q,\pi)=\pi.
    \end{equation*}
\end{lemma}

\begin{proof}
    This is a direct consequence of Lemma~\ref{lemma:zero-inverse-temp-value-policies}.
\end{proof}

\begin{lemma}\label{lemma:limiting-Gamma_2}
    Consider a sequence of inverse temperature parameterized by $\{\beta_t\}_{t=1}^\infty$ with $\beta_t > 0$, such that $\lim_{t \to \infty} \beta_t =0$.
    Then, for every Q-function $\Q\in Q$ and every prior pair $\rho \in \Pi$, the soft update operator $\Gamma_2^\beta$ converges pointwise to the standard update operator $\Gamma_1$. That is,
    \begin{align*}
        \lim_{t\to\infty}\Big[\Gamma_2^{\beta_t} (\Q,\rho)\Big](s,a\pl,a\op) &= \R(s,a\pl,a\op) +\gamma \sum_{s'} \T_{ss'}(a,\pl,a\op) \left[\bm{\rho}\pl(s')\trps \bm{\Q}(s') \bm{\rho}\op(s')\right] \\
        &= \left[\Gamma_1(\Q,\rho)\right](s,a\pl,a\op).
    \end{align*}
\end{lemma}

\begin{proof}
Note that
\begin{align}
\lim_{t\to\infty}&\Big[\Gamma_2^{\beta_t} (\Q,\rho)\Big](s,a\pl,a\op) = \lim_{\beta\to 0^+} \left[\Gamma^\beta_2 (\Q, \rho)\right] (s,a\pl,a\op) \nonumber \\
&=\R(s,a\pl,a\op)+ \gamma \sum_{s'\in\S}\T(s'|s,a\pl,a\op) \halfspace  \lim_{\beta\to 0^+} \Big\{\frac{1}{\beta\pl} \log \sum_{a\pl} \rho \pl(a\pl|s') \exp{ \beta\pl \Q^{\text{pl}}(s',a\pl)}\Big\} \label{eqn:lemm20-2}\\
&=\R(s,a\pl,a\op)+ \gamma \sum_{s'\in\S}\T(s'|s,a\pl,a\op) \halfspace  
\bigg[\sum_{a\pl, a\op} \rho\pl(a\pl|s') \rho\op(a\op|s') Q(s',a\pl,a\op)\bigg]\label{eqn:lemm20-3} \\
&= \left[\Gamma_1(\Q,\rho)\right](s,a\pl,a\op) \nonumber
\end{align}
The step from \eqref{eqn:lemm20-2} to \eqref{eqn:lemm20-3} utilized the result from Lemma \ref{lemma:zero-inverse-temp-value-policies}.
\end{proof}

\begin{theorem}\label{thm:limiting_Gamma_beta}
   Consider a sequence of inverse temperature parameterized by $\{\beta_t\}_{t=1}^\infty$ with $\beta_t > 0$, such that $\lim_{t \to \infty} \beta_t =0$.
    Then, 
    \begin{equation}
        \Gamma^0\bar{\Q} = \lim_{t \to \infty} \Gamma^{\beta_t} \bar{\Q} 
        = \lim_{\beta \to 0^+} \Gamma^\beta 
        \left[\begin{array}{c}
         \Q_{~}\\
         \Q_\KL 
    \end{array}\right]
    = \left[\begin{array}{c}
         \Gamma_1(\Q,\Gamma_\Nash(\Q))\\
         \Gamma_1(\Q_\KL,\Gamma_\Nash(\Q))
    \end{array}\right].
    \end{equation}
    Furthermore,
 $\Gamma^0$ is a strict contraction mapping with contraction factor $d$ defined in~\eqref{ddef}, and has $[\Q_\Nash^\star,\Q_\Nash^\star] \trps$ as its fixed point, where $\Q_\Nash^\star$ satisfies
    \begin{equation*}
        \Q_\Nash^\star = \Gamma_1(\Q_\Nash^\star, \Gamma_\Nash \Q_\Nash^\star).
    \end{equation*}
\end{theorem}

\begin{proof}
We first present the point-wise convergence of the operator $\Gamma^\beta$.
Consider the first component,
\begin{align}
    \lim_{t\to\infty} \Gamma_1(\Q,\Gamma_\KL^{\beta_t} (\Q_\KL, \Gamma_\Nash \Q))&=
    \lim_{\beta\to 0^+} \Gamma_1(\Q,\Gamma_\KL^\beta (\Q_\KL, \Gamma_\Nash \Q)) \\
    &=  \Gamma_1(\Q,\lim_{\beta\to 0^+}\Gamma_\KL^\beta (\Q_\KL, \Gamma_\Nash \Q)) 
    \label{appd-eqn:cor21a1}\\
    &= \Gamma_1(\Q,\Gamma_\Nash(\Q)) \label{appd-eqn:cor21a2}.
\end{align}
The step in \eqref{appd-eqn:cor21a1} is a result of the Lipschitz continuity of $\Gamma_1$ in $\Pi$ assumed in Assumption~\ref{assmpt:Gamma1-Reg},
and the step in \eqref{appd-eqn:cor21a2} is a result of Lemma \ref{lemma:limiting-Gamma_1}.

According to Lemma \ref{lemma:limiting-Gamma_2}, we have for the second component of $\Gamma^\beta$,
\begin{align*}
     \lim_{t\to\infty}\Gamma_2^{\beta_t}(\Q_\KL, \Gamma_\Nash \Q)=\lim_{\beta\to 0^+}\Gamma_2^\beta(\Q_\KL, \Gamma_\Nash \Q) = \Gamma_1(\Q_\KL,\Gamma_\Nash(\Q)).
\end{align*}

Since, for all $\beta$, $\Gamma^\beta$ is a contraction mapping with a contraction factor $d$, 
by the first result in Theorem \ref{thm:sequential-application-contraction}, the limiting operator $\Gamma^0$ as $\beta \to 0^+$ 
is also a contraction mapping with contraction factor $d$.
Thus, the point $[\Q_\Nash^\star,\Q_\Nash^\star] \trps$ is the unique fixed point of $\Gamma^0$.
\end{proof}

We now put the pieces together and prove the convergence of the baseline SNQ2 as in Theorem~\ref{thm:convergence-baseline-SNQ2}.


\begin{proof}(of Theorem~10)
    In Theorem \ref{thm:contraction_Gamma_beta}, we have shown in that under Assumption \ref{assmpt:Lip-Const-relationship}, the operators $\Gamma^{\beta_t}$ form a family of contraction mappings. 
    In Theorem \ref{thm:limiting_Gamma_beta}, we have shown that with a sequence  $\{{\beta_t}\}_{t=1}^\infty$ approaches 0, the operator $\Gamma^{\beta_t}$ converges pointwise to $\Gamma^0$, 
    which is a contraction mapping with $[\Q_\Nash^\star,\Q_\Nash^\star] \trps$ as its unique fixed point.
    Then, by Theorem~\ref{thm:sequential-application-contraction}, it follows that by sequentially applying $\Gamma^{\beta_t}$, the concatenated $Q$-function will converge to $[\Q_\Nash^\star,\Q_\Nash^\star] \trps$.
    By Shapley's Theorem~\cite{cMDP:2012}, the fixed point $\Q_\Nash^\star$ corresponds to the $Q$-function at the Nash equilibrium of the zero-sum stochastic game.
    Thus, the algorithm converges to a Nash equilibrium.
\end{proof}

\subsection{Convergence of the standard SNQ2 Algorithm}\label{subsec:conv-standard-SNQ2}

We extend the convergence proof on the baseline SNQ2 to the standard SNQ2, which updates the Nash policies every $M$ Q-function updates.


\begin{assumption} \label{assmpt:d_T}
Let the Lipschitz constants $d_i~i=1,\ldots,7$ of the operators $\Gamma_1, \Gamma_2^\beta, \Gamma_\Nash, \Gamma_\KL^\beta$ satisfy
    \begin{equation} \label{d_T_def}
       d_M = \max \Big\{d_\mathrm{in}^{\halfspace M} + d_5, (d_4+d_2 d_6)\Big(\sum_{k=0}^{M-1} d_\mathrm{in}^{\halfspace k}\Big)\Big\} < 1,
    \end{equation}
where $d_{\rm in} = \max \{d_1,d_2 d_7 + d_3\}$ and $M$ is the number of inner iterations.
\end{assumption}

\vspace{+5pt}
\begin{theorem}\label{thm:conv-standard-SNQ2}
 Under Assumption~\ref{assmpt:d_T}, the standard SNQ2 Algorithm converges to the Nash Q-function. 
\end{theorem}

\begin{proof}
    
Similarly to the proof of Theorem~\ref{thm:convergence-baseline-SNQ2}, we first show that the operation within the while loop in Algorithm~\ref{alg:cMDP-SNQ2-full} is a contraction mapping and then show the pointwise convergence and apply Theorem~\ref{thm:sequential-application-contraction}.
    
Define the operation conducted in the inner loop (lines 7 to 10) of Algorithm~\ref{alg:cMDP-SNQ2-full} as 
\begin{equation}
    \Gamma^{\beta,\pi_\Nash}_{\mathrm{in}} \bar{\Q} =
    \Gamma^{\beta,\pi_\Nash}_{\mathrm{in}}\left[ \begin{array}{c}
         \Q  \\
         \Q_\KL
    \end{array}\right] = \left[ \begin{array}{c}
         \Gamma_1(\Q,\Gamma_\KL^\beta(\Q_\KL, \pi_\Nash))  \\
         \Gamma_2^\beta(\Q_\KL,\pi_\Nash)
    \end{array}\right].
\end{equation}
Then, the result of the 
outer loop (lines 5 to 12) in Algorithm~\ref{alg:cMDP-SNQ2-full} can be captured through
\begin{equation}
    \Gamma^\beta_{\mathrm{out}} \left[ \begin{array}{c}
         \bar{\Q}  \\
         \pi_\Nash
    \end{array}\right] =
    \left[ \begin{array}{c}
         \Big( \Gamma^{\beta,\pi_\Nash}_{\mathrm{in}}  \Big)^M \bar{\Q} \\
         \Gamma_\Nash \Q
    \end{array}\right].
\end{equation}
The operator $\Gamma^\beta_{\mathrm{out}}$ operates on the space $\Q \times \Q \times \Pi$. 
We equip this space with the following metric
\begin{equation*}
    \widetilde{\D} \Bigg(\left[ \begin{array}{c}
         \bar{\Q}_1  \\
         \pi_{\Nash,1}
    \end{array}\right],
    \left[ \begin{array}{c}
         \bar{\Q}_2  \\
         \pi_{\Nash,2}
    \end{array}\right]\Bigg) = \D(\Q_1,\Q_2) + \D(\Q_{\KL,1},\Q_{\KL,2}) + \W(\pi_{\Nash,1},\pi_{\Nash,2}).
\end{equation*}

We now show the contraction factor of $\Gamma^{\beta,\pi_\Nash}_{\mathrm{in}} $ is bounded by 
\begin{equation}\label{appd-eqn:d_inner}
    d_\mathrm{in} = \max \{d_1,d_2 d_7 + d_3\}.
\end{equation}
Consider two concatenated Q-functions $\bar{\Q}_1, \bar{\Q}_2$, we have
\begin{align*}
    \bar{\D}(\Gamma^{\beta,\pi_\Nash}_{\mathrm{in}} \bar{\Q}_1,\Gamma^{\beta,\pi_\Nash}_{\mathrm{in}} \bar{\Q}_2) 
    & = \quad \D( \Gamma_1(\Q_1,\Gamma_\KL^\beta(\Q_{\KL,1}, \pi_\Nash),\Gamma_1(\Q_2,\Gamma_\KL^\beta(\Q_{\KL,2}, \pi_\Nash)) \\
    &\quad + \D(\Gamma_2^\beta(\Q_{\KL,1},\pi_\Nash),\Gamma_2^\beta(\Q_{\KL,1},\pi_\Nash)) \\
    &\leq d_1 \D(\Q_1,\Q_2) + d_1 d_7 \D(\Q_{\KL,1},\Q_{\KL,2}) + d_3 \D(\Q_{\KL,1},\Q_{\KL,2})\\
    &\leq \max \{d_1,d_2 d_7 + d_3\} \halfspace \bar{\D}(\bar{\Q}_1, \bar{\Q}_2)
    =d_\mathrm{in}\halfspace \bar{\D}(\bar{\Q}_1, \bar{\Q}_2).
\end{align*}

For the two Nash policies $\pi_{\Nash,1},\pi_{\Nash,2}$, we then have
\begin{align}
    \bar{\D}(\Gamma^{\beta,\pi_{\Nash,1}}_{\mathrm{in}} \bar{\Q},\halfspace \Gamma^{\beta,\pi_{\Nash,2}}_{\mathrm{in}} \bar{\Q}) 
    &=\quad \D( \Gamma_1(\Q,\Gamma_\KL^\beta(\Q_{\KL}, \pi_{\Nash,1}),\Gamma_1(\Q,\Gamma_\KL^\beta(\Q_{\KL}, \pi_{\Nash,1})) \nonumber \\
    &\quad + \D(\Gamma_2^\beta(\Q_{\KL},\pi_{\Nash,2}),\Gamma_2^\beta(\Q_{\KL},\pi_{\Nash,2})) \nonumber\\
    &\leq (d_2 d_6 + d_4) \W(\pi_{\Nash,1},\pi_{\Nash,2}). \label{appd-eqn:different_pi_nash}
\end{align}
\begin{equation}\label{eqn:induction-relationship}
    \bar{\D}\Big(\Big(\Gamma^{\beta,\pi_{\Nash,1}}_{\mathrm{in}}\Big)^{M} \bar{\Q},\halfspace \Big(\Gamma^{\beta,\pi_{\Nash,2}}_{\mathrm{in}}\Big)^{M} \bar{\Q}\Big) \leq \Big(\sum_{k=0}^{M-1}d_{\mathrm{in}}^{\halfspace k}\Big) \big(d_2 d_6+d_4\big) \halfspace \W(\pi_{\Nash,1},\pi_{\Nash,2}).
\end{equation}
The above result can be shown through induction. 
The case for $M=1$ is already shown in \eqref{appd-eqn:different_pi_nash}.
Now assume the relationship in \eqref{eqn:induction-relationship} holds for $M$, then for $M+1$, we have
\begin{align*}
    &\bar{\D}\Big(\Big(\Gamma^{\beta,\pi_{\Nash,1}}_{\mathrm{in}}\Big)^{M+1} \bar{\Q},\halfspace \Big(\Gamma^{\beta,\pi_{\Nash,2}}_{\mathrm{in}}\Big)^{M+1} \bar{\Q}\Big) \\
    &\leq \bar{\D}\Big(\Gamma^{\beta,\pi_{\Nash,1}}_{\mathrm{in}} \Big(\Gamma^{\beta,\pi_{\Nash,1}}_{\mathrm{in}}\Big)^{M} \bar{\Q},
    \halfspace \Gamma^{\beta,\pi_{\Nash,2}}_{\mathrm{in}}\Big(\Gamma^{\beta,\pi_{\Nash,1}}_{\mathrm{in}}\Big)^{M} \bar{\Q}\Big) \\
    &~~~~~~~+\bar{\D}\Big(\Gamma^{\beta,\pi_{\Nash,2}}_{\mathrm{in}} \Big(\Gamma^{\beta,\pi_{\Nash,1}}_{\mathrm{in}}\Big)^{M} \bar{\Q},
    \halfspace \Gamma^{\beta,\pi_{\Nash,2}}_{\mathrm{in}}\Big(\Gamma^{\beta,\pi_{\Nash,2}}_{\mathrm{in}}\Big)^{M} \bar{\Q}\Big) \\
    &\leq (d_2 d_6 + d_4) \W(\pi_{\Nash,1},\pi_{\Nash,2}) 
    + d_{\mathrm{in}}\halfspace \bar{\D}\Big(\Big(\Gamma^{\beta,\pi_{\Nash,1}}_{\mathrm{in}}\Big)^{M} \bar{\Q},\halfspace \Big(\Gamma^{\beta,\pi_{\Nash,2}}_{\mathrm{in}}\Big)^{M} \bar{\Q}\Big)\\
    & \leq (d_2 d_6 + d_4) \W(\pi_{\Nash,1},\pi_{\Nash,2}) + d_{\mathrm{in}} \Big(\sum_{k=0}^{M-1}d_{\mathrm{in}}^{\halfspace k}\Big) \big(d_2 d_6+d_4\big) \halfspace \W(\pi_{\Nash,1},\pi_{\Nash,2})\\
    &= \Big(\sum_{k=0}^{M}d_{\mathrm{in}}^{\halfspace k}\Big) \big(d_2 d_6+d_4\big) \halfspace \W(\pi_{\Nash,1},\pi_{\Nash,2}).
\end{align*}

Furthermore, for the outer loop operator, it follows
\begin{align}
    &\widetilde{\D}\Big(\Gamma^\beta_{\mathrm{out}} \left[ \begin{array}{c}
         \bar{\Q}_1  \\
         \pi_{\Nash,1}
    \end{array}\right],
    \Gamma^\beta_{\mathrm{out}} \left[ \begin{array}{c}
         \bar{\Q}_2  \\
         \pi_{\Nash,2}
    \end{array}\right]\Big) \nonumber\\
    &\leq
    \widetilde{\D}\Big(\Gamma^\beta_{\mathrm{out}} \left[ \begin{array}{c}
         \bar{\Q}_1  \\
         \pi_{\Nash,1}
    \end{array}\right],
    \Gamma^\beta_{\mathrm{out}} \left[ \begin{array}{c}
         \bar{\Q}_2  \\
         \pi_{\Nash,1}
    \end{array}\right]\Big)
    +
    \widetilde{\D}\Big(\Gamma^\beta_{\mathrm{out}} \left[ \begin{array}{c}
         \bar{\Q}_2  \\
         \pi_{\Nash,1}
    \end{array}\right],
    \Gamma^\beta_{\mathrm{out}} \left[ \begin{array}{c}
         \bar{\Q}_2  \\
         \pi_{\Nash,2}
    \end{array}\right]\Big)
    \nonumber \\
    &= \bar{\D}\Big( \Big(\Gamma^{\beta,\pi_{\Nash,1}}_{\mathrm{in}}\Big)^{M} \bar{\Q}_1,\Big(\Gamma^{\beta,\pi_{\Nash,1}}_{\mathrm{in}}\Big)^{M} \bar{\Q}_2\Big)  + \W(\Gamma_\Nash \Q_1, \Gamma_\Nash \Q_2) \label{eqn:134a}\\
    & \quad + \bar{\D}\Bigg(\Big(\Gamma^{\beta,\pi_{\Nash,1}}_{\mathrm{in}}\Big)^{M} \bar{\Q}_2,\Big(\Gamma^{\beta,\pi_{\Nash,2}}_{\mathrm{in}}\Big)^{M} \bar{\Q}_2\Big)  + \W(\Gamma_\Nash \Q_2, \Gamma_\Nash \Q_2)\label{eqn:134b}\\
    & \leq d_{\mathrm{in}}^{\halfspace T} \D(\Q_1,\Q_2) + d_{\mathrm{in}}^{\halfspace M} \D(\Q_{\KL,1},\Q_{\KL,2}) + d_5 \D(\Q_1,\Q_2)\label{eqn:134c}\\
    & + \Big(\sum_{k=0}^{M-1}d_{\mathrm{in}}^{\halfspace k}\Big) \big(d_2 d_6+d_4\big) \W(\pi_{\Nash,1},\pi_{\Nash,2})\label{eqn:134d}\\
    &\leq \max \Big\{d_\mathrm{in}^{\halfspace M} + d_5, (d_4+d_2 d_6)\Big(\sum_{k=0}^{M-1} d_\mathrm{in}^k\Big)\Big\} 
    ~ \D\Big(\left[ \begin{array}{c}
         \bar{\Q}_1  \\
         \pi_{\Nash,1}
    \end{array}\right],
    \left[ \begin{array}{c}
         \bar{\Q}_2  \\
         \pi_{\Nash,2}
    \end{array}\right]\Big).
\end{align}
From~\eqref{eqn:134a} to \eqref{eqn:134c}, we used the contraction factor in~\eqref{appd-eqn:d_inner}, and we used the result in \eqref{eqn:induction-relationship} to go from \eqref{eqn:134b} to \eqref{eqn:134d}.
In summary, we have shown that under Assumption~\ref{assmpt:d_T}, the operator $\Gamma^\beta_{\mathrm{out}}$ is a family of contraction mappings. 

Through similar arguments as in the proof of Theorem~\ref{thm:limiting_Gamma_beta}, one can show that the limiting operator $\Gamma^0_{\mathrm{out}}$ is a contraction and
the vector $[\Q_\Nash^\star,\Q_\Nash^\star, \pi_\Nash^\star] \trps$ is the fixed point of $\Gamma^0_{\mathrm{out}}$.
Then, per Theorem~\ref{thm:sequential-application-contraction}, it follows that the standard SNQ2 algorithm converges to a Nash equilibrium.

\end{proof}

\newpage
\bibliographystyle{ieeetr}
\bibliography{main.bib}

\end{appendices}

\end{document}